\documentclass[journal]{IEEEtran} 

\usepackage[colorlinks]{hyperref}
\usepackage{graphicx}
\usepackage{epstopdf}
\usepackage{amsmath}
\usepackage{float}
\usepackage{subfigure}
\usepackage{hyperref}
\usepackage{amssymb}
\usepackage{algorithm}
\usepackage{algorithmic}
\usepackage{units}
\usepackage{mathtools}
\usepackage{amsthm}
\newtheorem{theorem}{Theorem}
\usepackage{colortbl}
\usepackage{xcolor}

\begin{document}

\title{A Directional Antenna based Leader-Follower Relay System for End-to-End Robot Communications}

\author{Byung-Cheol Min$^{1}$, \and Ramviyas Parasuraman$^1$,  \and Sangjun Lee$^1$, \and Jin-Woo Jung$^{2,*}$, \and Eric T. Matson$^{1,3}$

\thanks{$^1$ Department of Computer and Information Technology, Purdue University, West Lafayette, IN 47907, USA.

$^2$ Department of Computer Science and Engineering, Dongguk University, Seoul, 04620, Republic of Korea.

$^3$ Department of Electronic Engineering, Kyung Hee University, Yongin, 17104, Republic of Korea.

$^*$ Corresponding author email: {\tt jwjung@dongguk.edu}.}

\thanks{This paper was presented in
part at the International Conference on Robot Intelligence Technology and Applications, Denver, USA, Dec, 2013 \cite{Min2014} and in part at the IEEE Sensors Applications Symposium, Queenstown, New Zealand, April, 2014 \cite{min2014using}.}
}
\markboth{Preprint version submitted to the Elsevier Journal of Robotics and Autonomous Systems}%
{Min \MakeLowercase{\textit{Min et al.}}: A Directional Antenna based Leader-Follower Relay System for End-to-End Robot Communications}

\maketitle
\begin{abstract}
In this paper, we present a directional antenna-based leader-follower robotic relay system capable of building end-to-end communication in complicated and dynamically changing environments. The proposed system consists of multiple networked robots - one is a mobile end node and the others are leaders or followers acting as radio relays. Every follower uses directional antennas to relay a communication radio and to estimate the location of the leader robot as a sensory device. For bearing estimation, we employ a weight centroid algorithm (WCA) and present a theoretical analysis of the use of WCA for this work. Using a robotic convoy method, we develop online, distributed control strategies that satisfy the scalability requirements of robotic network systems and enable cooperating robots to work independently. The performance of the proposed system is evaluated by conducting extensive real-world experiments that successfully build actual communication between two end nodes.

\end{abstract}

\begin{IEEEkeywords}
Multi-Robot System, Relay Robots, Robotic Convoy System, Wireless Communications, Directional Antennas, Weighted Centroid Algorithm.
\end{IEEEkeywords}

\IEEEpeerreviewmaketitle 

\section{Introduction}
\label{sec:intro}

\begin{figure}
\centering
\includegraphics[width=0.78\columnwidth, angle=90]{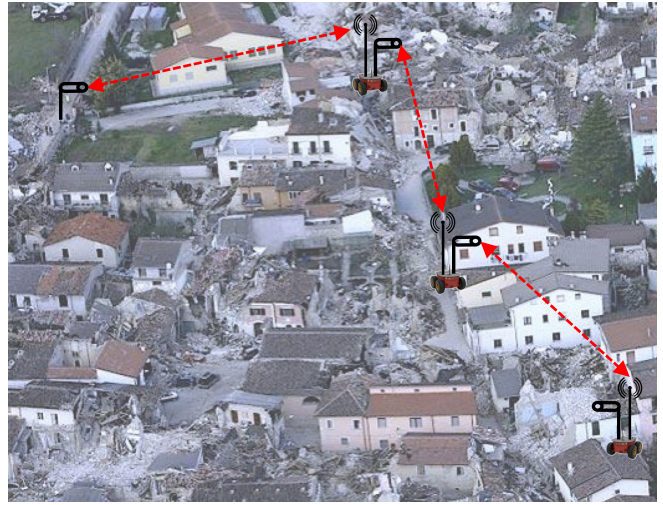}
\caption{A scenario of building the end-to-end communication link from a command center at the top left corner to the mobile end user or robot at the right bottom corner using relay robots in a disaster situation.}
\label{fig:1-1}   
\end{figure}

\IEEEPARstart{O}{ften}, long distance end-to-end networks need to be built on ad-hoc basis for a mobile end user (e.g., robots or first responders) in life-threatening and hazardous environments for applications in search, safety, security and rescue \cite{tuna_autonomous_2012,fischer2010location}. 
For instance, a firefighter exploring unstructured and unvisited places during {\color{black}an urban search and rescue} operation will need robust and long-range connectivity to a command center \cite{ko2009robot}, as illustrated in Figure \ref{fig:1-1}. 
A commonly proposed strategy to establish communication beyond the coverage area of a single radio is to deploy multiple mobile relay (or repeater) robots to relay wireless communication from the command center to the end user, building the end-to-end networks\footnote[1]{Throughout this paper, the term ``end-to-end network" or ``end-to-end communication" refers to the communications link between two end nodes, e.g., a command center and an end (lead or exploring) user.}\cite{nguyen_maintaining2_2004,tekdas2012building,tuna_autonomous_2012,parasuraman2014multi}. 
The mobility of the repeater robots will limit the mobility of the end users. Thus, to enable mobile end user movements in realizing assigned exploration tasks, the repeater robots should continuously adjust their positions accordingly to maintain end-to-end connectivity.

In this paper, we propose a leader-follower robotic relay system in order to build robust communication in complicated and dynamically changing environments. We will utilize a directional antenna based wireless device to realize long distance communication as well as a sensor to estimate the bearing (direction) of the leader or the precedent robot.

For bearing estimation, we employ a weight centroid algorithm (WCA). 
The proposed convoy system is online and uses distributed control strategy that satisfies the scalability requirements of robotic network systems and also allows cooperating robots to work independently.

In general, directional antennas provide a much better performance in terms of a communication range than omnidirectional antennas since directional antennas can concentrate radio signals in desired directions \cite{kranakis2004directional}, enhancing the range of our end-to-end communication. 
In our work, we will use a combination of omnidirectional and directional antennas so that our method is applicable to wide range of environments and in line-of-sight and non-line-of-sight conditions \cite{spyropoulos2003capacity}.
Directional antennas are a core component of the system that are {\color{black}readily available} off-the-shelf and cost-effective compared to other sensory devices such as laser range finder and vision system. 
In addition, WiFi technology will be used to establish a wireless local area network (LAN), which allows high speed data transmission through the entire end-to-end network. This capability allows our convoy system to transfer and receive large amounts of data for real-time video broadcasting from a point of interest.

In our prior works, we introduced a novel bearing estimation algorithm {\color{black} using WCA and signal strength measurements from a directional antenna in \cite{min2013design}}. In \cite{Min2014,min2014using}, we proposed a leader-follower system using the bearing estimation system, and demonstrated the network throughput performance of the system with experiments involving two robots. We also integrated the convoy scheme with obstacle avoidance algorithm in \cite{Min2014}. We further extended the convoy system with thorough establishment of the WCA based bearing estimation algorithm and demonstrated the performance of the convoy system using multi-robot experiments in \cite{Min_JFR}. We also compared various combinations of antennas (omnidirectional-omnidirectional, omnidirectional-directional, directional-directional) of the leader and follower robots in \cite{Min_JFR}, and showed the advantages of using a combination of using omnidirectional and directional antennas for sensing the direction of arrival {\color{black}(DOA)} and throughput improvements in different environments. Moreover, in \cite{Min_JFR}, we assumed the antennas are static in the analysis of WCA. 

Having provided a solid foundation for robotic convoy system using directional antennas, in this paper, we extend our prior works and contribute in the following ways.
\begin{itemize}
\item We thoroughly verify and analyze the WCA algorithm for bearing estimation and consider {\color{black}the Doppler effect} due to the movement of high speed antenna tracking system and the robot. 
\item We show how the bearing estimation algorithm integrates with the obstacle avoidance system and mobile robot control and prove convergence in convoying operation.
\item We conduct extensive experiments in both indoors and outdoors to validate the proposed convoy system using end-to-end throughput tests and examine the effects of relay robots and the number of hopping. 
\end{itemize}

The paper is organized as follows. The next section presents relevant works in building end-to-end robot communications, and highlights novel aspects of the proposed approach. In Section \ref{sec:background}, we provide background information and briefly present the problem statement. We then describe our approach for building an end-to-end network using a robotic convoy system in Section \ref{sec:approach}. In Section \ref{sec:experiments}, a set of experiments on our methods and systems that demonstrate our solutions to the end-to-end communication problem are included. Section \ref{sec:conclusions} summarizes the conclusions and future scope of this work.

\section{Related Work}
\label{sec:relatedwork}

The task of building end-to-end communication using multiple relay robots can be approached based on the type of end node - a static end node or a dynamic end node \cite{Min201623}. 
As shown in \cite{tekdas2012building}, for static end nodes, end-to-end communication can be realized by planning relay robots' paths to their final positions prior to deployment. However, it is often challenged because a predefined path is not robust to unknown and dynamic environments. 
On the other hand, for dynamic end nodes, deploying a group of leader-follower robots is studied in \cite{Nguyen_autonomous_2002, nguyen_maintaining2_2004}. This approach takes advantages of multiple pairs of mobile robots to build a communication network. It could be realized by having the front-most robot be the mobile end node and making the follower robots simply chase the leader or precedent robots without making any decisions \cite{Hogg_sensors_2001}. This feature makes the group of leader-follow robots as a convoy team, enabling the system to be robust to dynamic environments and easy to extend the range. The main challenge here is that the performance of the entire system is highly dependent on the accuracy of on-board sensors used to chase the leader robots. Thus, many researches focus on control algorithms with various types of sensors in order to improve the performance of follower robots \cite{5766050, hong2008distributed, chen2009leader}. Some research focused on statistical models of error sources that can cause degradation of navigation performances \cite{6182732, patwari2003relative}, while connectivity control has been studied in \cite{zavlanos2011graph,tardioli2016robot} to maintain communication between multiple connected robots.

Typically, follower robots with one or multi on-board sensors are used to implement leader-follower strategies. Vision-based leader-follower systems have been widely used in convoy applications \cite{Giesbrecht_a_2009, Mariottini_vision-based_2009, Cowan_vision-based_2003, Goi_vision-based_2010}. By estimating the leader's position from the sequence of a video image, the follower is able to follow the leader \cite{min2009vision, borenstein_human_2010}. 
Although this approach has been demonstrated in practice, there is an inherent constraint that the leader must stay in the follower's vision. 
In real life, loss in vision sensing is shown to happen frequently \cite{jia2009vision}, thus alternative compensation strategies need to be incorporated. 

Instead of vision sensors, inertial sensors including GPS (Global Positioning System) have been employed to estimate the leader's position and heading in \cite{4671092, 5175357, petrov2009nonlinear}. During motion, the follower robots follow the GPS paths provided by the leader robot. 
As this approach does not require line-of-sight condition, it is known to be a good alternative to vision-based systems. However, GPS signal is sensitive to location and the environment, and hence prone to blind spots, e.g., in indoor environments. Thus such sensing scheme may result in estimation errors, and can cause the system to fail. Also, such system require constant data exchanges between the leader and the follower robots. 

\begin{figure*}[!h]
\centering
\includegraphics[width=1.9\columnwidth]{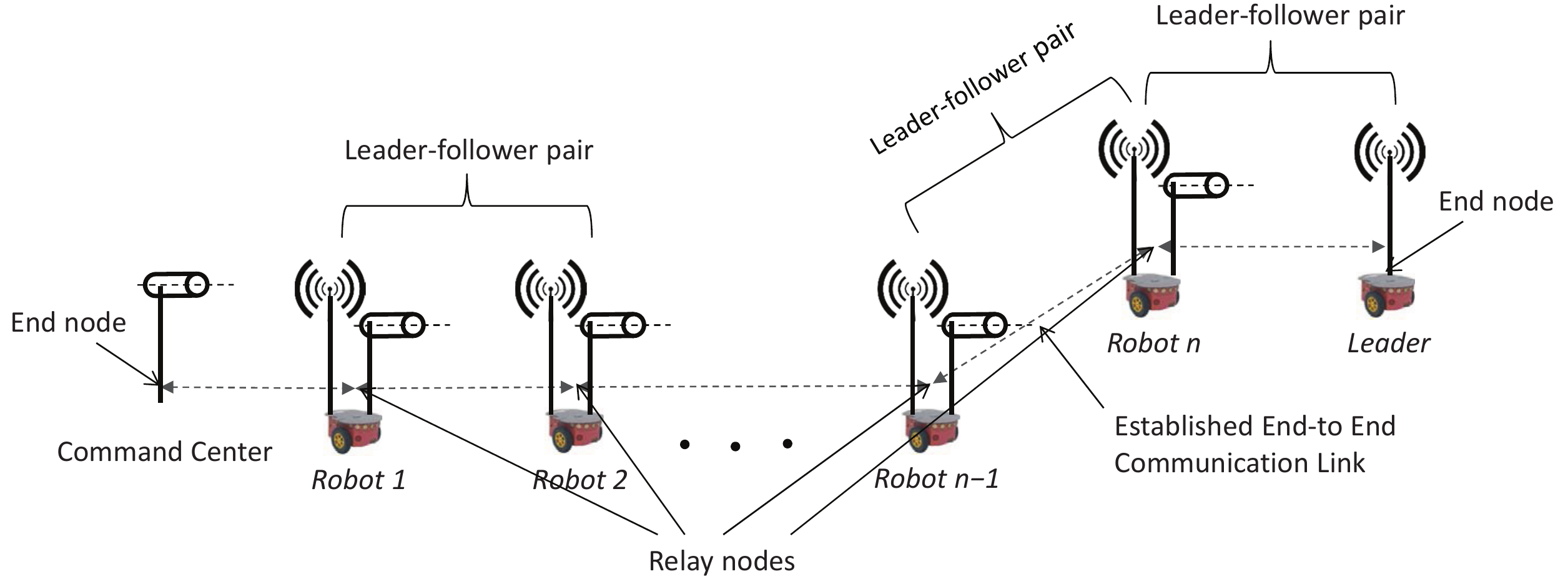}
\caption{Overview of a team of leader-follower robots creating end-to-end communication links.}
\label{fig:6-1}
\end{figure*}
In this paper, we estimate the bearing of the leader robot through active antenna tracking and processing of the wireless signal strength, instead of relying on vision or inertial based sensors. In literature, multiple directional antennas have been used to estimate bearing of radio signals \cite{Bezzo2014,caccamo2015extending}, however they increase power consumption and resource utilization of the robots, although proving advantages such as fast processing and added redundancy.

We depart from previous works in the following ways.
We use directional antennas to measure signal strength and to communicate with the leader robot, and thus the system can cope with cases where the sight of its leader is lost and enable safe navigation. In principle, as long as the antenna is pointing in the direction in which the leader robot is moving, the system works efficiently. Since we use active tracking with a motorized pan tilt system, we are able to keep track of the leader robot bearing and are able to recover from cases where it loses the direction by constantly scanning for signals around the follower robot.
We also present how we realize the entire system with the {\color{black}commercial off-the-shelf devices} and conduct extensive real-world experiments. The proposed system is highly beneficial for cooperative multi-robot systems in dynamic unstructured environments where it is critical to rapidly make adjustments to facilitate communication.

\section{Background}
\label{sec:background}
In this section, we first present how radio signal strength is modeled. Then we continue to describe the problem we tackle and introduce the leader-follower system.

\subsection{Radio signal strength}
Radio signal strength has been widely used in robotics literature for predictions and improvements in wireless communication performance \cite{tardioli2010enforcing,parasuraman2014wireless}. 
{\color{black}The received signal strength (RSS) is given by} \cite{Fink_online_2010,lindhe2013exploiting}:

\begin{equation}
{P_{dBm}} = {L_0} - \underbrace { 10n \cdot \log \left( {\left\| {{x^t} - \left. x \right\|} \right.} \right)}_{Path \, Loss} - \underbrace {f\left( {{x^t},x} \right)}_{Shadowing} - \underbrace \varepsilon _{multipath}
\label{eq:6-13}
\end{equation}

\noindent where $L_0$ is the measured (reference) power at 1 meter from the transmitter, $n$ is the decay exponent, and $x^t$ and $x$ are the positions of the transmitter and receiver respectively. 
While the path loss fading due to distance is deterministic, the shadowing and multipath fading are often modeled as stochastic processes with (zero-mean) Gaussian and Nakagami distribution respectively.
Precisely modeling the shadowing and multipath effects are difficult \cite{lindhe2013exploiting, Bruggemann2009}, and often not required in the proposed application as they are relatively smaller in effects compared to fading. Nevertheless, in our bearing estimation algorithm, the multipath effects are filtered out due to (spatial) averaging of the signal values over the angular spectrum.

\subsection{Problem statement}
\label{sec:problem}
The main challenge is to develop online distributed control strategies that satisfy the scalability requirements of robotic network systems and that enable cooperating robots to work independently to create end-to-end communication in dynamic and unknown environments. Estimation and tracking of bearing using directional antennas is the core of the proposed convoy system. A robot convoy team is composed into identical leader-follower pairs, as shown in Figure \ref{fig:6-1} \cite{Goi_vision-based_2010}. The lead robot, on the right of the figure, is a mobile end node and guides a team in a robot convoy. This end node can also be a human user carrying/wearing the node. The goal of a follower robot is to autonomously track the trajectory of its immediate leader and then relay a radio signal between the leader in front and the robot behind.

\subsection{System Description}
\label{ch:6-2}

\begin{figure}
\centering 
\includegraphics[width=0.95\columnwidth]{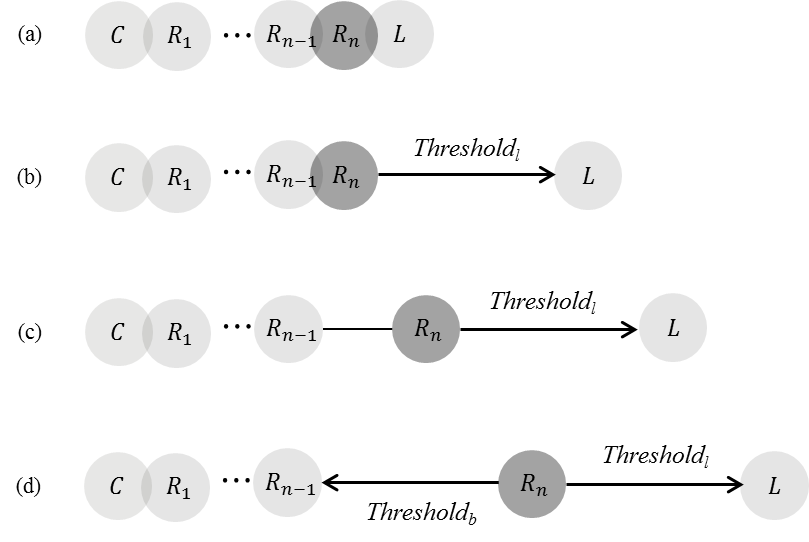} 
\caption{Visual description of our convoy strategy, focusing on the follower case with $R_n$: (a) Initial state, (b) $drive$ is activated, (c) $R_n$ in $drive$ tracks the team leader $L$, (d) $R_n$ stops tracking because $stop$ is activated.} 
\label{fig:6-2}
\end{figure}

An overview of the strategy for a robotic convoy is depicted in Figure \ref{fig:6-2}. Let us assume there are $n$ networked robots and are associated to the command center sequentially. Each robot is responsible for measuring the Received Signal  Strength Indicator (RSSI) of the robot preceding it as well as the robot behind it, in order to monitor wireless connectivity between the two neighboring nodes. More specifically, in order for a robot in the convoy to maintain high quality of wireless connection with its leader, the robot must track its leader if the RSSI connection with the leader has reached the assigned RSSI threshold; that is, the condition if ${RSSI^{l} \le Threshold_l}$ then {\em drive} is activated. This case is depicted in Figure \ref{fig:6-2}(b) with a focus on $R_{n}$. The status of the robot in $drive$, tracking its leader, is presented in Figure \ref{fig:6-2}(c). Similarly, in order for a robot to maintain good wireless connection with the robot behind it, the robot must stop tracking its leader if the RSSI connection with the robot behind it has reached the assigned RSSI threshold; that is, the condition if ${RSSI^b \le Threshold_b}$ then {\em stop} is activated. This case is depicted in Figure \ref{fig:6-2}(d) with a focus on $R_n$. 
This achieves our goal to extend end-to-end communication in a convoy fashion.

\begin{figure*}
\centering
\includegraphics[width=0.95\textwidth]{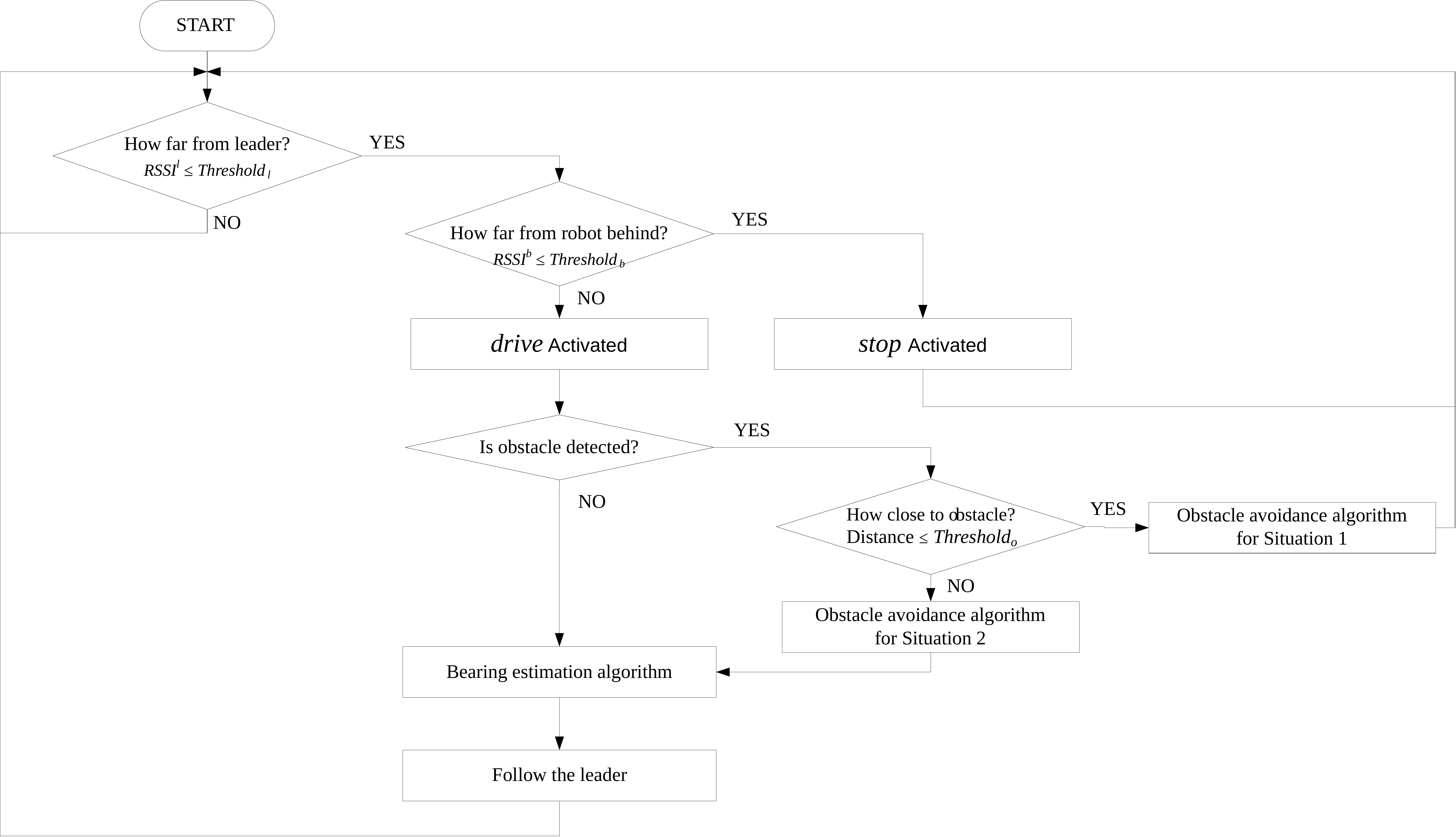}
\caption{A flow chart of the follower robotic system.}
\label{fig:6-3}       
\end{figure*}

For every robot in the convoy to be autonomous, each robot operates with the predefined steps, depicted in Figure \ref{fig:6-3}. Note that the leader robot at the end node runs separately. 
In Figure \ref{fig:6-3}, the robot in the convoy is composed mainly of two algorithms $-$ a bearing estimation algorithm and an obstacle avoidance algorithm. The bearing estimation algorithm allows a follower robot to track its leader's trajectory. For bearing estimation, we employ the WCA that was proposed in \cite{Min_JFR}. The obstacle avoidance algorithm allows the follower robot to avoid the obstacle between itself and the leader. For the obstacle avoidance algorithm, we consider two different situations as described in \cite{Min2014}, depending on the distance between an object and the follower robot. 

In Situation 1,  if the distance is too close (i.e., when the measured distance is smaller than the pre-defined threshold, $Threshold_{o}$), obstacle avoidance becomes a top priority, i.e., the robot stops following the leader and avoids the obstacle and tries to avoid a crash. For this approach, a simple obstacle avoidance algorithm was developed and more details will be described in Section \ref{ch:6-5}. 
In Situation 2, if the distance is not close, but an object is detected on the path, both a bearing estimation algorithm and obstacle avoidance algorithm are concurrently initiated, i.e., the robot keeps tracking the leader while avoiding the object. 

\begin{figure}[t]
\centering
\includegraphics[width=0.95\columnwidth]{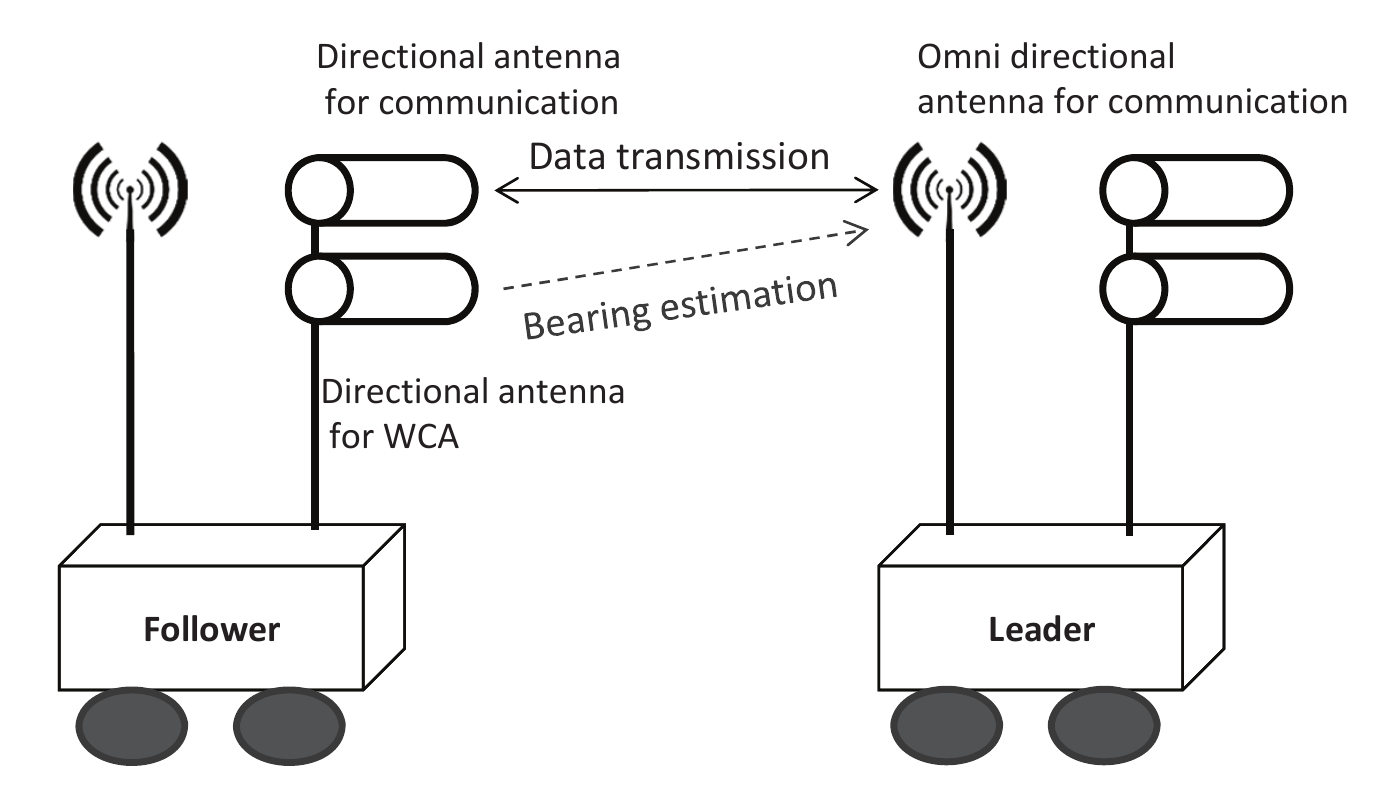}
\caption{Directional sensing model for leader-follower robotic system.}
\label{fig:6-4} 
\end{figure}

In the robotic system, a network device with antennas is installed on each robot for two purposes - one is to maintain wireless connectivity between the two end nodes, and the other is to provide and measure a wireless signal for bearing estimation. Specifically, {\color{black}each robot is equipped with an omni-directional antenna} and two directional antennas as shown in Figure \ref{fig:6-4}. The follower estimates the bearing to the transmitter on the leader by measuring its radio power using the bottom directional antenna and by utilizing the estimated bearing to track the trajectory of the leader robot. The bearing estimate also allows computing of the best orientation for the top directional antenna that is associated with the omni-directional antenna on the leader for actual data transmission in end-to-end communication. With the assumption that the two communication sides are far enough apart, the two antennas on the follower are installed on the same vertical axis which enables that the field of views from each antenna are projected onto a single space. 

\section{Robotic Convoy System}\label{sec:approach}
In this section, we first briefly describe the WCA for bearing estimation of the follower robot. Then, we validate the algorithm against {\color{black}the Doppler effect} due to high speed rotations in antennas. Analysis of the algorithm for convoying operation is then provided, followed by the integration of WCA with the robot control and obstacle avoidance.

\subsection{Bearing Estimation Algorithm}
\begin{figure}[t]
\centering
\includegraphics[width=0.95\columnwidth]{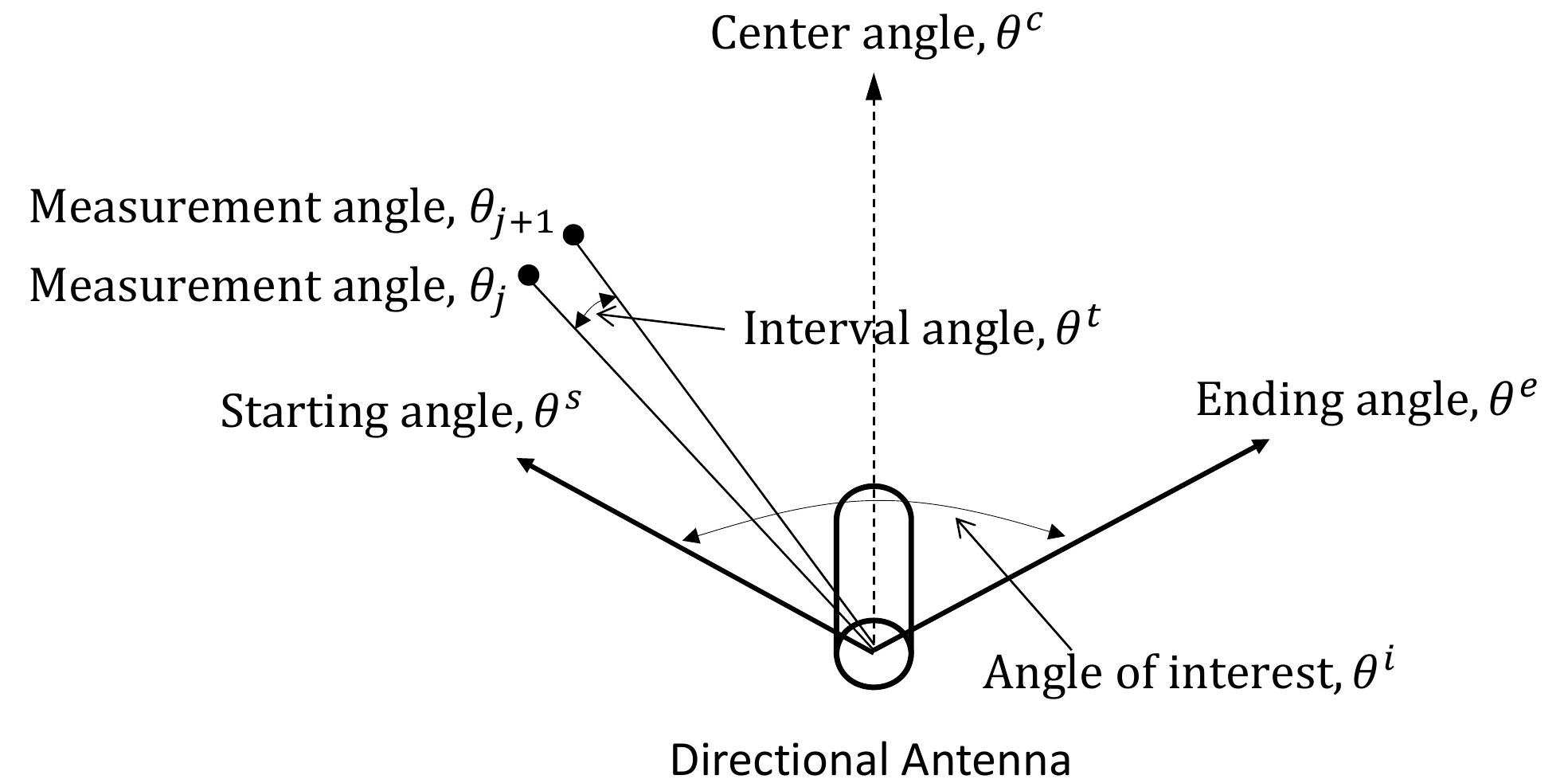}
\caption{Defined parameters for bearing estimation with a directional antenna when rotating to clockwise.}
\label{rita4}
\end{figure}

In Figure \ref{fig:6-4}, the bottom directional antenna on the follower robot that can rotate horizontally with the help of a servo motor, is used for bearing estimation. The parameters of bearing estimation with a directional antenna are described in Figure \ref{rita4}. Note that the interval angle $\theta^t$ can be computed by dividing the interesting range by the total number of measurements $N_t$.

In the first step of the WCA, a single rotary directional antenna measures $N_t$ signal strength while it rotates at a constant angular velocity $\dot{\theta}$ from  $\theta^s$ to $\theta^e$ and produces a set of ${RSSI}_j$ values, where $j$ is the index of the measurement such that  $j \in \left\{ {1,2, \ldots ,N_t} \right\}$. Here, $\theta^s$ and $\theta^e$ are determined by the center angle $\theta^c$ that is set to be aligned with the previous bearing to prevent the estimation from approaching the end where an actual bearing dwells \cite{Min_JFR}.

In the second step, the relationship between a weight and the signal strength measurements at $\theta_j$ is defined as:
\begin{equation}
w_j\ = 10^{\left( \frac{RSSI_{j}}{\gamma} \right)}
\label{eq:2}
\end{equation}
where $\gamma$ is the positive gain that should be appropriately determined in every application scenario. The equation implies that a higher RSSI value has more weight than a lower RSSI value. Then, the bearing can be estimated by means of weighted centroid approaches as:

\begin{equation}\label{ritaeq2}
{\rm{\widetilde \Theta}} = \frac{\sum\nolimits_{j = 1}^{{N_t}} {w_j\theta_j}}{\sum\nolimits_{j = 1}^{{N_t}} {w_j}}.
\end{equation}

\subsection{Assumptions} 

In our robotic convoy system, multiple pairs of leader-follower mobile robots build the connection from end to end nodes. A single pair of leader-follower robots is comprised of a leader robot that moves towards a desired location or chases its leader robot in another pair, and a follower robot that chases the leader robot, in order to maintain the connection alive. This implies that the role of each robot is decided to be a leader or a follower depending upon its current goal, and each pair of leader-follower robots performs independently as shown in Figure \ref{fig:6-1}. In this way, without loss of generality, the convergence analysis of the WCA for the entire system can be addressed by showing the proof of the case where a single pair of robots is considered. 
The following assumptions are considered:
	\begin{itemize}
		\item[$A1$] An unconstrained model is considered. A leader robot can freely move in a given area.
		\item[$A2$] A constant velocity is considered. 
	\end{itemize}

The second assumption means that there is no change in the velocities of the leader robot and the follower robot during a scanning of the directional antenna on the follower robot. Their velocities are updated only when the current scanning is done and right before the next scanning is performed. This assumes that the change in the robot's velocity is only affected by $k$, not by $j$ with the $j_\text{th}$ measurement angle at the $k_\text{th}$ scanning of the directional antenna. This is to simplify the mathematic modeling and is valid if the rotation speed of {\color{black}the} directional antenna is faster than the velocity of the mobile robot. 

\subsection{Doppler Effect}
\label{sec:doppler}
Due to the movement of the directional antennas, both the self rotation and the movement of robots, it is imperative to analyze {\color{black}the impact of the Doppler effect} on the bearing estimation performance. 
The Doppler effect is the change in frequency of a wave for an observer moving relative to its source. It is observed whenever the source of waves is moving with respect to an observer. It is presumed that the robotic convoy system is not completely free from the Doppler effect since the source of the radio signals (antennas) on both the leader and follower robots are mobile \cite{Bok2014}.

A typical power equation at the receiver in dB takes the following form:
	\begin{equation}\label{eq:power}
	P_{r}=P_{t}+G_{t}+G_{r} \cos^{2}{\phi} -L_{f}
	\end{equation}
where $P_{t}$ is the output power of the transmitting antenna; $G_{t}$ is the gain of the transmitting antenna; {\color{black}$G_{r}$ is the gain of the receiving antenna;} $L_{f}$ is the free-space loss given by $20\log_{10}(4\pi d f / c)$ where $d$ is the distance between the antennas, $f$ is the frequency of the radio signal, {\color{black}and $c$ is the velocity of waves in the medium}. The gain at the receiving antenna {\color{black}$G_{r}$ depends} on the angular separation $\phi$ between the transmitter and the receiver antennas. Note that $L_{f}$ in (\ref{eq:power}) could vary if the Doppler effect takes place due to the change in frequency caused by relative motion between source and destination nodes. 

The relationship between observed frequency $f^*$ and source frequency $f$ for our system is given by:
	\begin{equation}\label{doppler}
	f^*=f \left( \frac{ c \pm V_{\text{follower}}}{1 \pm V_{\text{leader}}} \right) = f \left( 1 \pm \frac{V_{\text{relative}}}{c} \right)
	\end{equation}
where $V_{\text{follower}}$ is the velocity of the follower and is positive if it is moving towards the leader; $V_{\text{leader}}$ is the velocity of the leader and is positive if it is moving away from the follower. $V_{\text{relative}}$ is the relative velocity between the leader and follower robots. Note the scanning antennas do not circularly polarize the light waves and hence do not create any Doppler shift, however may affect the performance of antenna tracking which will be discussed in Section~\ref{sec:tracking}. 

\begin{figure}[t]
\centering
\includegraphics[width=0.75\columnwidth,angle=90]{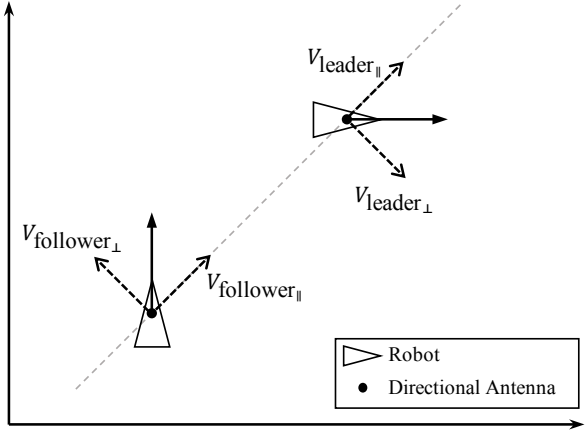}
\caption{A pair of leader-follower robots and its velocity components. Each of perpendicular and parallel component can be found in (\ref{v_com_per}) and (\ref{v_com_pa}).}
\label{fig:v} 
\end{figure}

The velocity vector of a robot can be separated into parallel and perpendicular components to an inertial frame of reference as shown in Figure \ref{fig:v}, and they are determined by:
	\begin{align}
	V_{\text{robot}_{\parallel}} &= V_{\text{robot}} \cos{\big( \theta_{\text{cen}}(k)-\theta_{\text{cen}}^{\ast}(k) \big)} \label{v_com_per}\\
	V_{\text{robot}_{\perp}} &= V_{\text{robot}} \sin{\big( \theta_{\text{cen}}(k)-\theta_{\text{cen}}^{\ast}(k) \big)}\label{v_com_pa}
	\end{align}
where $\theta_{\text{cen}}^{\ast}(k)$ is the actual DOA trajectory that the center angle should be placed at each scanning and the subscript robot represents both a leader and a follower. 

The free-space loss in \ref{eq:power} becomes,
	\begin{equation}
	L_{f^*} = 20 log_{10} (d) + 20 log_{10} (\frac{4\pi f }{c}) + 20 log_{10} (\frac{c \pm V_{\text{relative}}}{c}).
    \label{eq:loss}
	\end{equation}

{\color{black}This equation indicates that} the relative motion between robots affects the distance between them during a scan, which will result in the path loss measurements impacted by velocity other than directionality of the directional antenna alone. However, using Assumption A2, we can assert that the relative velocity $V_{\text{relative}}$ between the robots does not change within a scanning period. {\color{black}Therefore, the Doppler effect during a scanning period that is led by a rotating directional antenna for the estimation of relative bearings in our robotic convoy system does not exist.}

Moreover, the RSSI fluctuations due to shadowing during a scan become nullified because of the same weight factor applied across the range $\theta^s$ to $\theta^e$. Therefore, the WCA is unaffected by the changes in instantaneous path loss measurements. {\color{black}In addition}, the WCA acts as a filter in itself, thus it is robust to multipath effects.

\subsection{Convergence}
\label{sec:convergence}
In mathematics, the term ``convergence" is generally defined as a property of approaching a limit more and more closely as a variable of the function increases or decreases or as the number of terms of the series increases. The limit could be a constant or a function. 
Recall that we assume the leader robot is freely moving ({\color{black}Assumption} A1).
{\color{black}This indicates that the following robot cannot fundamentally and mathematically converge until the leader robot is fixed to a certain position or a certain direction. Thus, the essential process becomes the tracking and converging problem where the follow robot chases the moving leader robot (called ``tracking mode" in the next subsection \ref{sec:tracking}) and finds the bearing of the leader robot when the leader robot stops moving (called ``converging mode" in this subsection).} That is, {\color{black}in the tracking mode,} the directional antenna of the follower robot should keep a position of the leader robot within {\color{black}the angle of interest} $\theta^i$ with a consideration of the potential movement range of the leader robot, in order to not lose the leader robot and to continuously track it while the leader robot is moving.

In the tracking and converging process, the converging mode is active when both the leader robot and the follower robot are stationary. {\color{black}Therefore, as shown in the previous subsection \ref{sec:doppler}, the Doppler effect does not take place in the converging mode. Given that, we can assume $P_{t}$, $G_{t}$, and $L_{f}$ in (\ref{eq:power}) are constant, and} the subsequent center angle $\theta^c$ of the directional antenna by the WCA can be obtained by \cite{Min_JFR}:
	\begin{align}\label{centerangle}
	\begin{split}
	{\theta _{cen}}(k + 1) = {\theta _{cen}}(k) \\ 
	&\hspace*{-45pt}+ \frac{{{\theta _{int}}}}{{2N}}\frac{{\sum\nolimits_{j =  - N}^N {j{{10}^{\frac{{{G_r}{{\cos }^2}\left( {{\theta _{cen}}(k) - {\theta _{cen}}^ * + \frac{{j{\theta _{int}}}}{{2N}}} \right)}}{\gamma }}}} }}{{\sum\nolimits_{j =  - N}^N {{{10}^{\frac{{{G_r}{{\cos }^2}\left( {{\theta _{cen}}(k) - {\theta _{cen}}^ * + \frac{{j{\theta _{int}}}}{{2N}}} \right)}}{\gamma }}}} }}
	\end{split}
	\end{align}

\begin{equation}
e(k + 1) = e(k) + \frac{{{\theta _{int}}}}{{2N}}\frac{{\sum\nolimits_{j =  - N}^N {j{{10}^{\frac{{{G_r}{{\cos }^2}\left( {e(k) + \frac{{j{\theta _{int}}}}{{2N}}} \right)}}{\gamma }}}} }}{{\sum\nolimits_{j =  - N}^N {{{10}^{\frac{{{G_r}{{\cos }^2}\left( {e(k) + \frac{{j{\theta _{int}}}}{{2N}}} \right)}}{\gamma }}}} }}
\end{equation}
where $e(k)={\theta _{cen}}(k) - {\theta _{cen}}^ *$, i.e., $e(k)$ is the error between the estimated DOA at the $k_\text{th}$ scanning and the actual DOA.

If we define $f\left( {e(k)} \right)$ as the differentiation of $e(k)$ in discrete time, then 
	\begin{equation}
	f\left( {e(k)} \right) = \frac{{{\theta _{int}}}}{{2N}}\frac{{\sum\nolimits_{j =  - N}^N {j{{10}^{\frac{{{G_r}{{\cos }^2}\left( {e(k) + \frac{{j{\theta _{int}}}}{{2N}}} \right)}}{\gamma }}}} }}{{\sum\nolimits_{j =  - N}^N {{{10}^{\frac{{{G_r}{{\cos }^2}\left( {e(k) + \frac{{j{\theta _{int}}}}{{2N}}} \right)}}{\gamma }}}} }}.
	\end{equation}

\begin{figure}
  \centering 
  \subfigure[$\gamma=10$]{\label{fig:4-15-a}  \includegraphics[width=0.45\textwidth ]{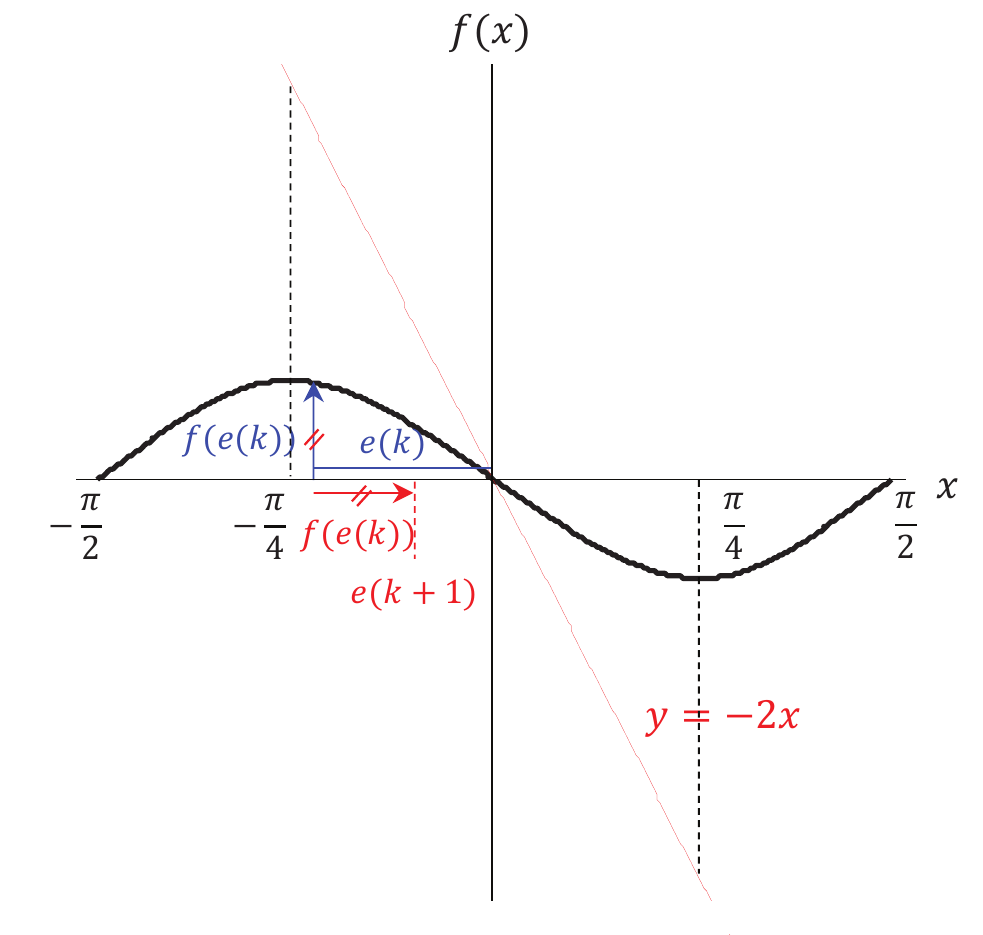}}
  \hskip 0.1truein
  \subfigure[$\gamma=1$]{\label{fig:4-15-b}\includegraphics[width=0.45\textwidth]{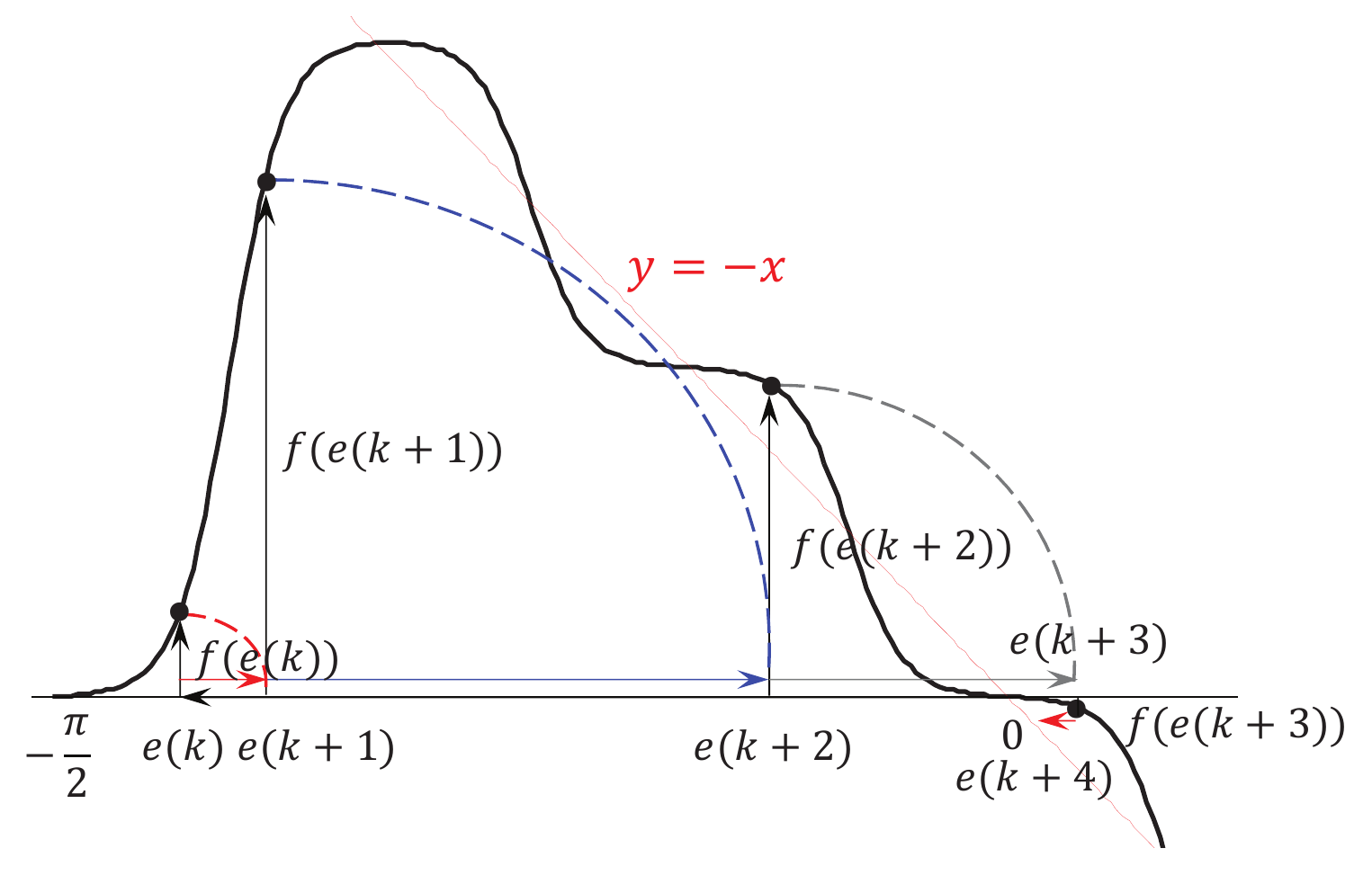}}
  \caption{Verification of convergence of the modified WCA. (a) shows that the absolute value of $f(x)$ is always less than the absolute value of $y=-2x$. Therefore,  as $k$ increases, $f(x)$ goes to zero. (b) shows an enlarged graph with a setting of $\gamma$ to 1 for more detailed explanations of the verification \cite{Min_JFR}.}
  \label{fig:wca_graphic}    
\end{figure}

As illustrated in Figure \ref{fig:wca_graphic}, the absolute error at the $(k+1)$ scanning is always less than the absolute error at the $k_\text{th}$ scanning, because  $f\left( {e(k)} \right) > 0$ in the negative domain $e(k) \in \left[ { - \frac{\pi }{2},0} \right]$, $f\left( {e(k)} \right) < 0$ in the positive domain $e(k) \in \left[ { 0,\frac{\pi }{2}} \right]$, and $f\left( {e(k)} \right) = 0$ at $e(k)=0$. Moreover, as shown in Figure \ref{fig:4-15-a}, the absolute value of $f(x)$ is always less than the absolute value of  $y=-2x$. If it was greater than the absolute value of $y=-2x$, a divergence would take place.
    
\subsection{Tracking}
\label{sec:tracking}
Under the condition that the follower robot could continuously track the leader robot without loosing it, the convoying objective is not compromised. This objective can be achieved by limiting the relative velocity using Theorem \ref{Theorem:tracking}. 

\begin{theorem}\label{Theorem:tracking}
Assume that the angular velocity (scanning) of the directional antenna is $\dot{\theta}$, the instantaneous distance between the leader robot and the following robot is ${d}$, and the maximum scanning angle is $\theta_{max}$. The DOA can be tracked under one of the following two conditions: 
\begin{enumerate}
\item $V_{\text{relative}} \leq \frac{\dot{\theta} d}{\theta_{max}}$
\item $V_{\text{leader}_{\perp}} \leq ( \dot{\theta} d + \theta_{max} V_{\text{relative}} )$
\end{enumerate}
\end{theorem}

	\begin{figure}[t]
	\centering
	\includegraphics[width=0.45\columnwidth,angle=90]{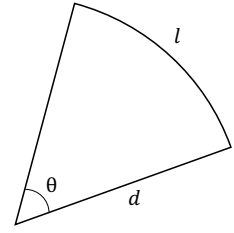}
	\caption{Circle sector enclosed by two radii and an arc.}
	\label{arc} 
	\end{figure}
    
\begin{proof}
Considering a circular sector as shown in Figure \ref{arc}, the general relationship between arc length and its partial angle $\theta$ (angle between the antennas of leader and follower) is represented as:
	\begin{equation}
	l = d\theta,
	\end{equation}	
and its partial derivative is given by:
	\begin{equation}
	\frac{\partial l}{\partial t} = \frac{\partial d}{\partial t} \theta + d\frac{\partial \theta}{\partial t}. \label{eq:partial}
	\end{equation}	

Tracking is feasible if the arc length does not change between scans. i.e., if $\frac{\partial l}{\partial t} \cong 0$. A change in the distance is given by $V_{\text{relative}}$, and a change in the $\theta$ is given by scanning velocity $\dot{\theta}$. Assuming the scan angle is {\color{black}within} the scanning range $\theta \leq \theta_{max}$, the  equation becomes:
	\begin{equation}
	V_{\text{relative}} \leq \frac{ \dot{\theta} d} {\theta_{max}}.
	\end{equation}

The perpendicular velocity of the leader robot $V_{\text{leader}_{\perp}}$ (perpendicular to the relative motion of leader-follower) should be within the arc length. Thus, $V_{\text{leader}_{\perp}}$ is limited by Eq.~\ref{eq:partial} {\color{black}as follows}:
	\begin{equation}\label{theta_approx}
	V_{\text{leader} \perp} \leq \frac{\partial l}{\partial t} \leq ( \dot{\theta} d + \theta_{max} V_{\text{relative}} ).
	\end{equation}

This concludes the proof.
\end{proof}

This implies that the tracking is feasible and the WCA based tracking is unaffected by both the movement of the robots and antennas as long as the velocities are limited depending on the scanning speed, range and the leader-follower separation. 

{\color{black}In the tracking mode, the Doppler effect could have an impact because $L_f$ can be varied according to the relative velocity between the leader robot and follower robot. Thus, the power in (\ref{eq:power}) and the next center angle in (\ref{centerangle}) are then rewritten as:
	\begin{align}
	\begin{split}
	P_{r} = P_{t}+G_{t} +G_{r} \cos^{2}{\left( \theta_{j}(k) - \theta_{cen}^{\ast}(k)\right)} - L_{f*}(k)
	\end{split}
	\end{align}

	\begin{align}
	\begin{split}
	{\theta _{cen}}(k + 1) = {\theta _{cen}}(k) \\ 
	&\hspace*{-65pt}+ \frac{{{\theta _{int}}}}{{2N}}\frac{{\sum\nolimits_{j =  - N}^N {j{{10}^{\frac{{{G_r}{{\cos }^2}\left( {{\theta _{cen}}(k) - {\theta_{cen}}^* (k)  + \frac{{j{\theta _{int}}}}{{2N}}} \right)}-L_{f*}(k)}{\gamma }}}} }}{{\sum\nolimits_{j =  - N}^N {{{10}^{\frac{{{G_r}{{\cos }^2}\left( {{\theta _{cen}}(k) - {\theta _{cen}}^* (k)  + \frac{{j{\theta _{int}}}}{{2N}}} \right)}-L_{f*}(k)}{\gamma }}}} }}.
	\end{split}
	\end{align}

Such changes indicate that a weight represented as 10 power exponent in the WCA could be changed by the pointing error angle of the directional antenna and the relative velocity between the leader robot and the follower robot. Nonetheless, the changes are represented as another form of weight and becomes nullified because of the same weight factor applied across the range $\theta^s$ to $\theta^e$. Moreover, it is clear from Eq.~\ref{eq:loss} that the relative velocity  $V_{\text{relative}}$ has a negligible effect on the path loss equation ($c>>V_{\text{relative}}$). (e.g. the robots velocities are 1 m/s while the velocity of light is 3$\cdot10^8$~m/s). This implies that the Doppler effect over WCA based tracking is completely insignificant.}

\subsection{Mobile Robot Control}
\label{ch:6-5}
The P3-AT is a four-wheeled robot as shown in Figure \ref{fig:6-9}, however, two wheels on the same side are physically interconnected with a rubber belt. For the simple control of this robot for the robotic convoy system, differential-drive mobile robots with characteristics of non-slipping and pure rolling are considered. The robot can be then controlled to move to any posture by adjusting the velocity of the left wheel $V_L$ and the velocity of the right wheel $V_R$:

\begin{equation}
\begin{array}{l}
{V_L} = {v} + {k_{{p}}} {\rm{\widetilde \Theta}}  + {k_{{d}}}({\rm{\widetilde \Theta}}  - {\rm{\widetilde \Theta} ^{\it{t} - \rm{1}}})\\
{V_R} = {v} - {k_{{p}}} {\rm{\widetilde \Theta}}  - {k_{{d}}}({\rm{\widetilde \Theta}}  - {\rm{\widetilde \Theta} ^{\it{t} - \rm{1}}}).
\end{array}
\label{eq:6-9}
\end{equation}

In this {\color{black}paper}, by definition a follower may face two different obstacle situations $-$ {\color{black}Situation 1} is when an obstacle is too close (e.g., the distance between the robot and the obstacle is less than 1 meter), and {\color{black}Situation 2} is when an obstacle is detected, but is not close. It is worth noting that we briefly described these two situations in \ref{ch:6-2}. 

In {\color{black}Situation 1}, since there is a high chance that a collision can take place, the robot should stop tracking the leader, and first avoid the object by utilizing a set of sonar sensors. Therefore, a simple obstacle avoidance algorithm was developed \cite{Min2014} that is also based on a weighted-centroid approach where a weight is computed from the measured sonar distances, calculating a direction to guide the robot to a safe region {\color{black}using the following expression:

\begin{equation}
{\rm{\widetilde \Theta}}  = \frac{{\sum\nolimits_{k = 1}^{{N_s}} {{w_k}{\phi _k}} }}{{\sum\nolimits_{k = 1}^{{N_s}} {{w_k}} }}
\label{eq:17} 
\end{equation}

\noindent where $N_s$ is the total number of sonar measurements, and $w_k\ = {10^{\left( {\frac{{{-Distance}}_k}{{\gamma}}} \right)}}$, where $\gamma$ is a positive gain, $Distance_k$ is a measured sonar distance at $\phi_k$, and $k$ $\in \left\{ {1,2, \ldots ,N_s} \right\}$. This algorithm is mainly designed to avoid an obstacle, so it should be activated only when there is an object detected by a sonar sensor and the measured distance is shorter than a pre-determined threshold, ${Threshold}_o$. Therefore, if this algorithm is activated, then ${\rm{\widetilde \Theta}}$ in (\ref{eq:17}) is used as the direction that guides the robot to a safe region in (\ref{eq:6-9}).}

The algorithm for Situation 1 was developed to prevent the follower robot from colliding with objects. With this algorithm, we can prevent most of the crashes into obstacles. However, if we take only this situation into consideration, the motions of the robot may become too large upon approaching an obstacle. For this reason, we developed another algorithm \cite{Min2014} for dealing with {\color{black}Situation 2} where an obstacle is detected, but is not close. Implementing this second algorithm helps in reducing the chances that the robot will face a dangerous situation from close objects such as in {\color{black}Situation 1}. By changing heading in advance upon approaching close objects, the robot is able to more effectively and safely follow the leader.

\begin{figure}[t]
{\color{black}
\centering
\includegraphics[width=0.95\columnwidth]{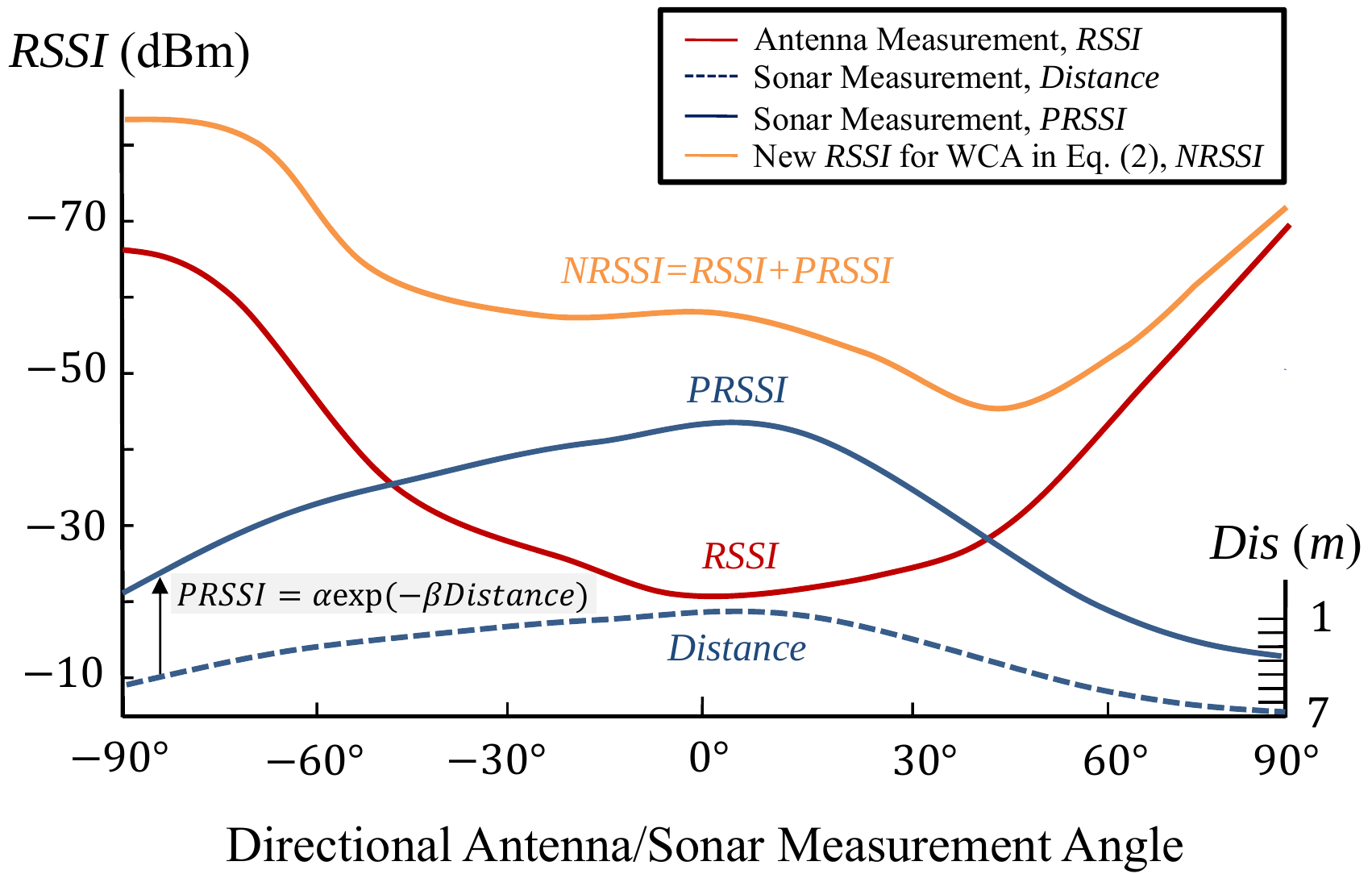}
\caption{A concept of the penalty function. A pseudo RSSI measurement is generated with sonar measurement and integrated to a real RSSI measurement.}
\label{fig:prssi}   
}
\end{figure}
\begin{figure*}[!h] 
{\color{black}
\centering
\includegraphics[width=1.4\columnwidth]{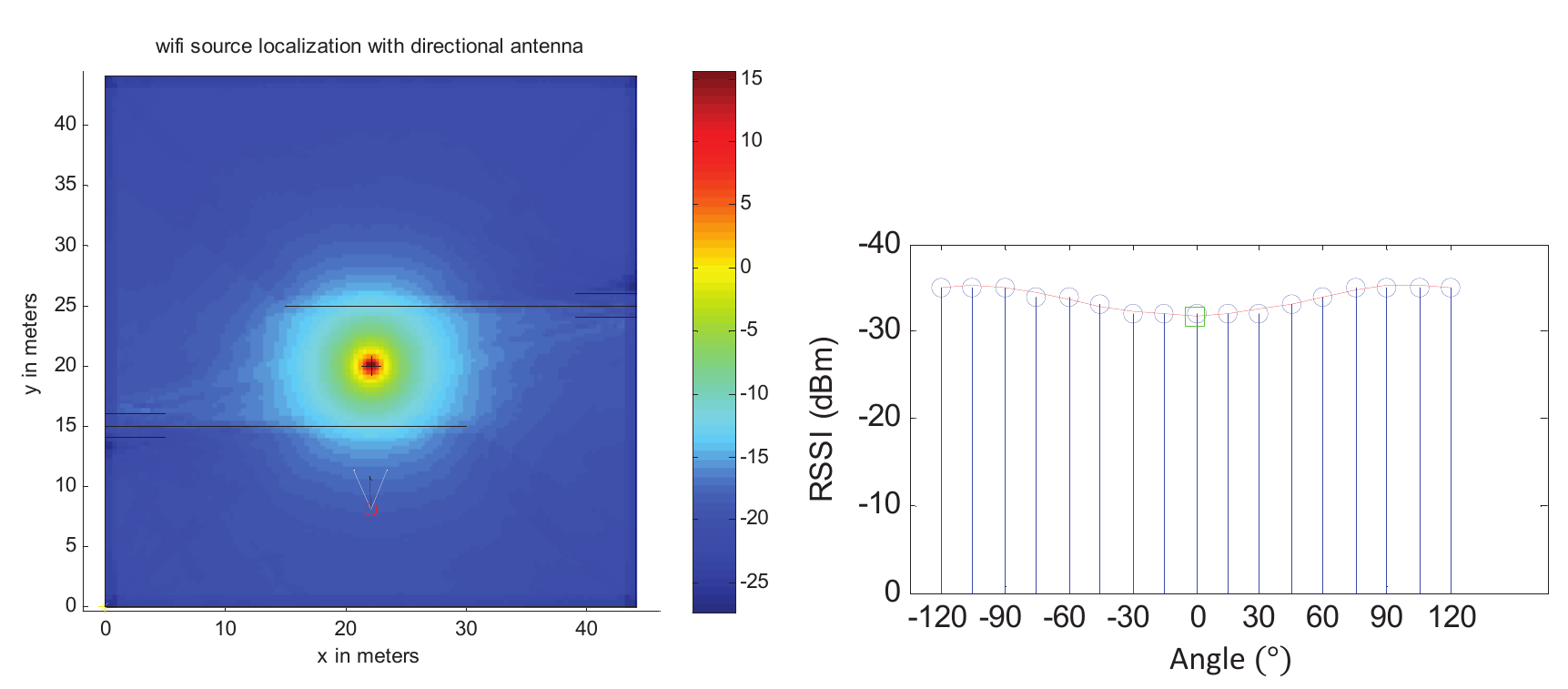}
\vspace{-5pt}
\caption{Bearing estimation only with real RSSI measurement. The left figure shows that the bearing was estimated to around $0^\circ$ (see the black arrow), making the follower keep moving toward the front wall, and the right figure shows the measured RSSI.}
\label{fig:wo_penalty}   

\vspace{15pt}
\centering 
\includegraphics[width=1.4\columnwidth]{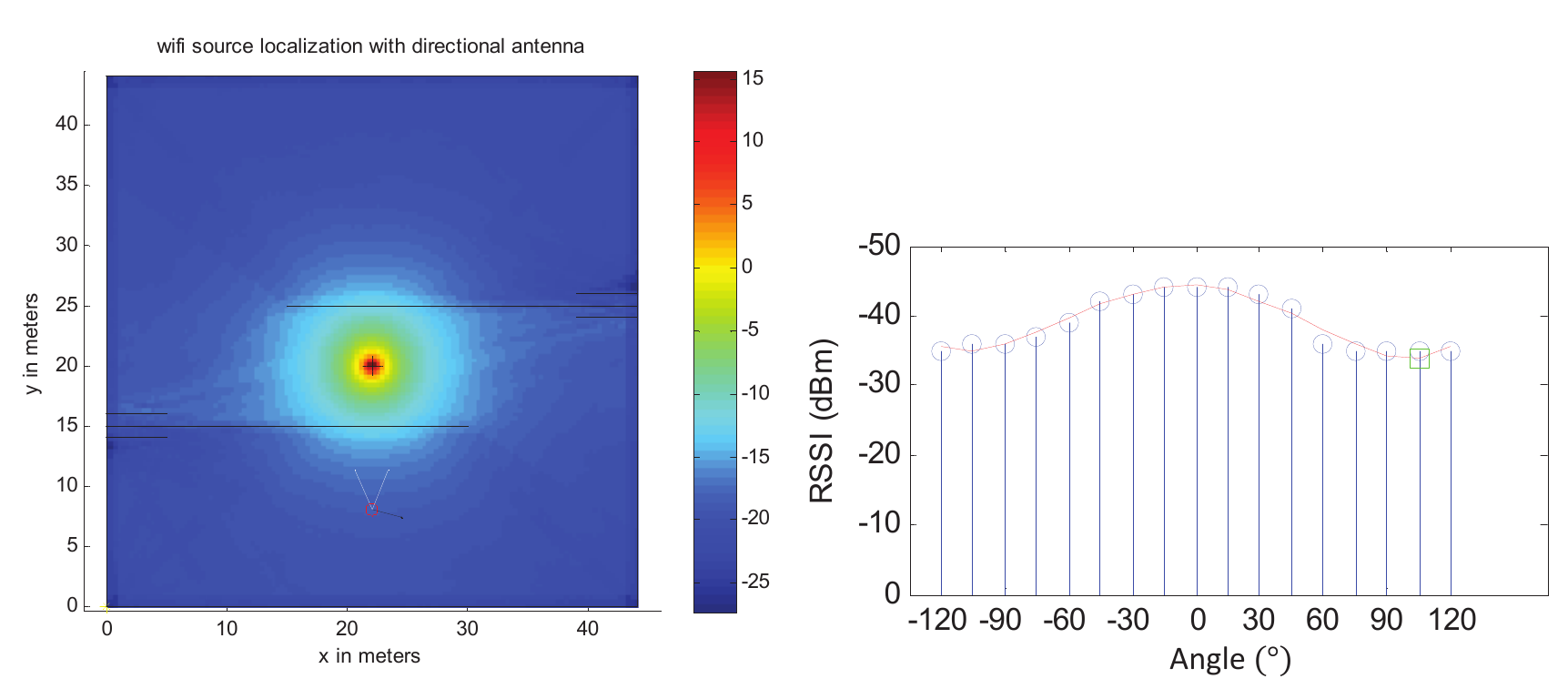}
\vspace{-5pt}
\caption{Bearing estimation with real RSSI measurement, with a penalty levied. The left figure shows that the bearing was estimated to over $+90^\circ$ (see the black arrow), enabling the follower to take an action to avoid crashing into the wall in advance while keeping tracking its leader, and the right figure shows the measured RSSI with pseudo RSSI added. A video demonstrating this obstacle avoidance algorithm in MATLAB simulation can be found at \href{https://youtu.be/o9tTFi_PkLI}{https://youtu.be/o9tTFi\_PkLI}.}
\label{fig:with_penalty}  
}
\end{figure*}

Sonar sensors are utilized again for this algorithm in the form of a penalty function. {\color{black}The basic concept of the penalty function is to integrate a sonar sensor measurement into an antenna measurement as depicted in Figure \ref{fig:prssi}. If an object on the path is detected, then the function generates a pseudo RSSI measurement that is factored into a real RSSI measurement, producing a new RSSI value for WCA in (\ref{eq:2}), denoted with $NRSSI$. The pseudo RSSI is generated by:
\begin{equation}
PRSSI_k=\alpha\text{exp}(-\beta Distance_k)
\label{eq:18}
\end{equation}

\noindent where $k \in \left\{1,2, \ldots ,N_s \right\}$, and $\alpha$ and $\beta$ are constants for regulating the level of the penalty function. These two parameters should be carefully determined, depending on the material of the object that is detected by the sonar sensors. For example, if the material of the object is impenetrable to a wireless signal, $\alpha$ could be set to a lower value. However, if the material of the object is penetrable, $\alpha$ should be set high enough for the obstacle to be recognized.	Details about the effect on varying $\alpha$ can be found in \cite{Min2014}.

Figure \ref{fig:wo_penalty} and Figure \ref{fig:with_penalty} show the effectiveness of the penalty function. Figure \ref{fig:wo_penalty} depicts a case where there are only real RSSI measurements, and the bearing was estimated to around $0^\circ$ when the robot is in Situation 2. In this case, the robot would keep moving toward the front wall with the estimated bearing and suddenly change its heading as soon as the robot lies in Situation 1 (e.g., when the distance between the robot and the obstacle is less than 1 meter). In this situation, the robot would be able to avoid a collision, but not be able to keep tracking its leader anymore. On the other hand, Figure \ref{fig:with_penalty} depicts a case where pseudo RSSI measurements are levied as a penalty to real RSSI measurements. In this case, the bearing was estimated to over $+90^\circ$, pointing toward a roadway that enables the follower to take an action to avoid crashing into the wall in advance while keeping tracking its leader.} 

This penalty function is activated only when there is an object detected by a sonar sensor, and its measured distance is longer than the pre-determined threshold for {\color{black}Situation 1}. Therefore, if the follower lies in {\color{black}Situation 2} or in which the robot is free of obstacles and can keep tracking the leader, it runs with the bearing estimation algorithm, activating (\ref{eq:6-9}) for velocity control. Therefore, $\rm{\widetilde \Theta}$ is the current estimated bearing obtained in (\ref{ritaeq2}), $\rm{\widetilde \Theta} ^{\it{t} - \rm{1}}$ is the old estimated bearing, $k_{p}$ and $k_{d}$ are positive gains, and $v$ is the background velocity of the robot, set to change according to a value of the best RSSI measurement from one scan, i.e., $v$ is calculated by
\begin{equation}
{v} =  - {\omega _1}RSS{I^l} - {\omega _2}
\label{eq:6-11}
\end{equation}

\noindent where ${RSSI}^l$ indicates the best RSSI measurement in one scan, $\omega_1$ and $\omega_2$ are should be set to a positive value and ${w_2} \le \left| {{w_1} \cdot RSS{I^l}} \right|$ for $v$ to be a positive value.

For the robot stopping criteria, we use the following conditions,
\begin{equation}
\left\{ \begin{array}{ll}
{V_L}~\text{and}~{V_R} = 0 &\text{if}~{{RSSI}^l} \le Threshold_{l}~ \\
& \text{and}~{{RSSI}^b} \ge Threshold_{b}\\
{V_L}~\text{and}~{V_R}~\text{from (\ref{eq:6-9})} ~~&\text{else}.
\end{array} \right.
\label{eq:6-12}
\end{equation}

\noindent In (\ref{eq:6-12}), ${{RSSI}^b}$ indicates the RSSI measurement from the node behind the robot. Depending on a value of ${Threshold}_l$, we can differentiate how close the follower can get to the leader or prevent the follower from getting too close to the leader. Also, a value of ${Threshold}_b$ determines how far the follower can drive away from the robot behind it.

Then, we can roughly calculate $P_{dBm}$ by pre-obtaining $L_0$ and $n$ with experiments. Therefore, we can select a proper value of ${Threshold}_l$ and ${Threshold}_b$ with (\ref{eq:6-13}) for the desired motion of our follower system. For example, we identified through experiments that $-15$ dBm of ${Threshold}_l$ keeps the follower away from the leader at intervals of 1 meter in indoor environments and $-20$ dBm for outdoor environments.


\section{Experiments}\label{sec:experiments}
\subsection{Experimental Setup}
\label{ch:6-6-1}

\begin{figure}
  \centering 
  \subfigure[A complete robotic convoy team]{\label{fig:6-9-a}  \includegraphics[width=0.9\columnwidth]{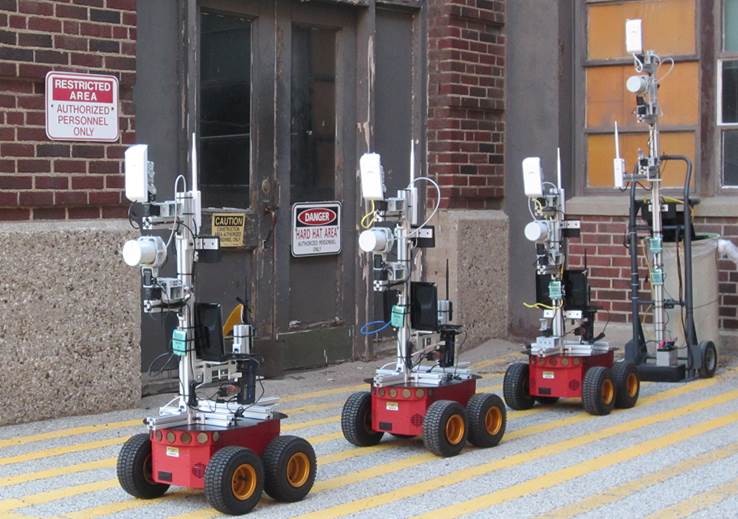}}\\
  \subfigure[A robot with network components]{\label{fig:6-9-b}\includegraphics[width=0.95\columnwidth]{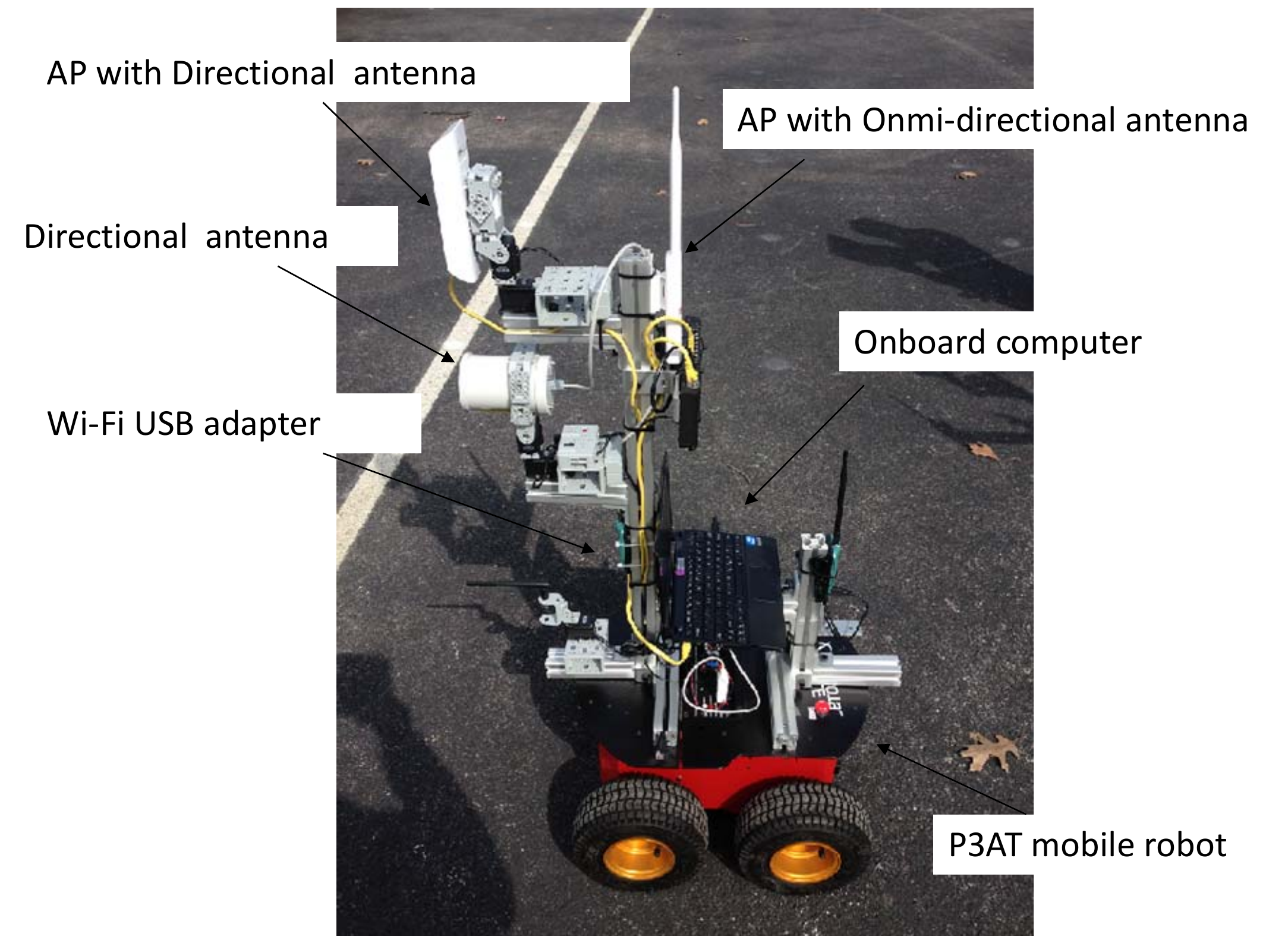}}
  \caption{Robotic convoy system. A robot was built with the commercial off-the-shelf network devices and robot parts and accessories.}
  \label{fig:6-9}
\end{figure}

A prototype of the robotic convoy system was developed and is shown in Figure \ref{fig:6-9}. Each robot is homogeneous and equipped with the same capabilities. Each robotic system is made up of the P3-AT mobile robot, a laptop, AP (Access Point) with an omni-directional antenna, AP with a directional antenna, a directional antenna with a Wi-Fi USB adapter, a network switch and two pan-tilt servo devices, as shown in Figure \ref{fig:6-9-b}.

For bearing estimation, we installed a small and light Yagi antenna, manufactured by PCTEL. For the top directional antenna on a follower, we installed $Nano$ $Station loco M2$, manufactured by Ubiquiti. For the transmitter on the leader, requiring an omni-directional antenna, we used a  $PicoStation M2HP$. 

The P3-AT, pan-tilt devices, and Wi-Fi USB adapter are connected by a serial connection to the laptop that processes all required algorithms and methods. A pan-tilt device allows the directional antenna to be autonomously oriented to a specific angle. In this {\color{black}paper}, we employ a pan angle only as the directional antennas we chose for this project have about 55$^\circ$ beamwidth vertically, and therefore there are few cases where our robots would be deployed out of the range vertically. Nonetheless, it is worth noting that vertical beamwidth would also affect wireless communication in some cases.

\begin{figure*}
  \centering 
\includegraphics[width=1.9\columnwidth]{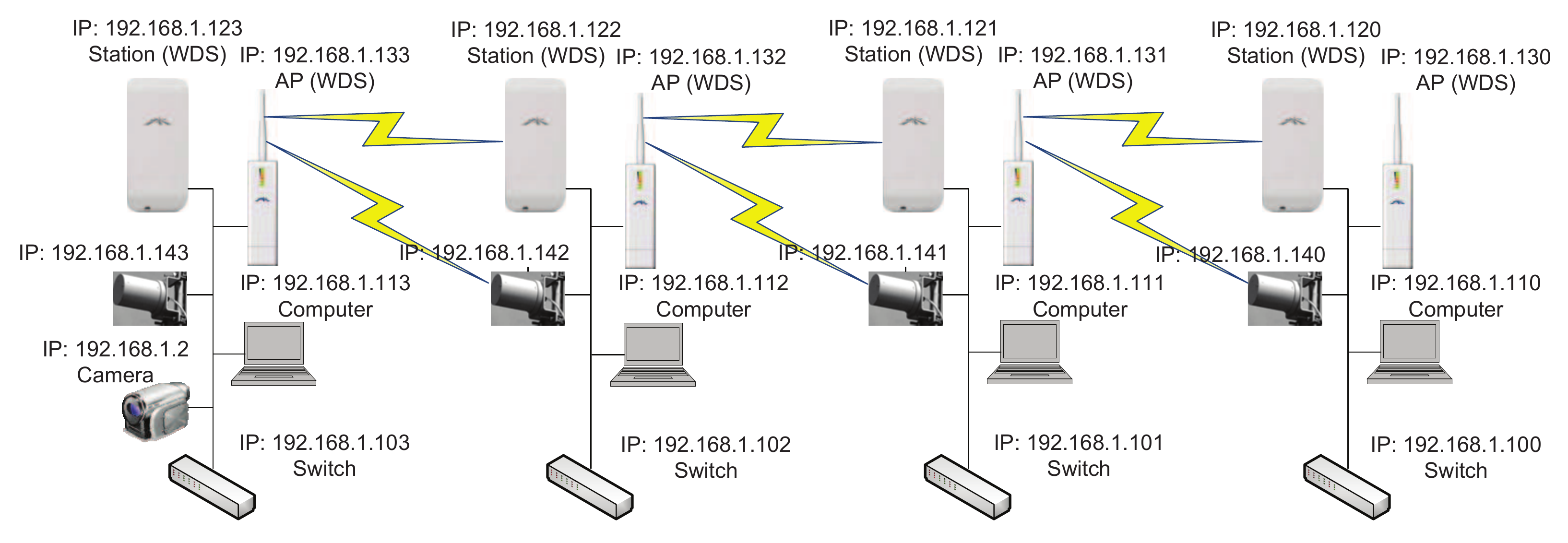}
  \caption{A configuration of robotic broadband networks.}
  \label{fig:6-network}
\end{figure*}

For the network configuration, we set up the indirect point-to-point link with off-the-shelf network devices as depicted in Figure \ref{fig:6-network}. This configuration acts like a very long wired cable and allows us to build a transparent ethernet bridge between two end nodes wirelessly. The left most node is one of the end nodes, having a network camera, that in this {\color{black}paper} could be a robot or a human user. The right most node is the other end node, i.e., that is a command center. Every device including antennas, computers, and switches has a static IP address as shown in Figure \ref{fig:6-network} for remote access. For example, accessing ``192.168.1.2" allows us to watch real-time video from the camera, and accessing ``192.168.1.112" allows remote control of the follower robot behind the end node.

The parameters needed in the WCA, the obstacle avoidance algorithm, and robot control were set and are set as follows: $\theta^i = 180^\circ, {Threshold}_o = 800$ cm, $\gamma = 10, k_{p} = 1.0, k_{d} = 0.3, w_1 = 10, w_2 = 150$. Due to the physical limitation of servo motors in our pan-tilt system, we set $\theta^i$ to be $180^\circ$. This setting results in the initial scan performed at $\theta^s$ = $-90^\circ$, $\theta^e$ = $90^\circ$. $N_t$ was approximately 25 for most of the tests. These settings were applied to all of the tests. We varied ${Threshold}_l$ and ${Threshold}_b$ according to test purposes. They are presented later along with each test.

\begin{figure} 
  \centering 
  \subfigure[A helmet network device]{\label{fig:6-11-a}  \includegraphics[width=0.33\textwidth]{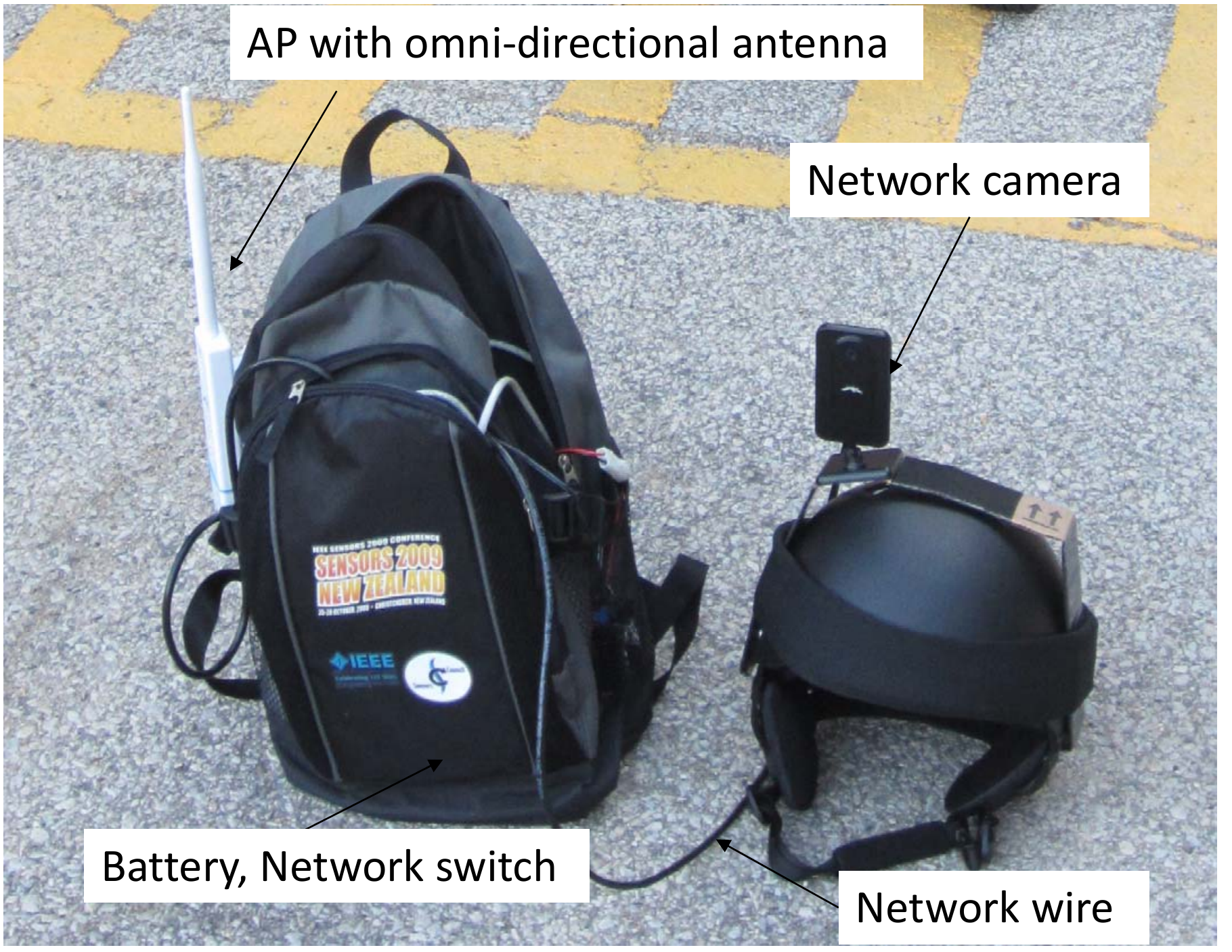}}
 \hskip 0.08truein
  \subfigure[A human user carrying the device]{\label{fig:6-11-b}\includegraphics[width=0.1335\textwidth]{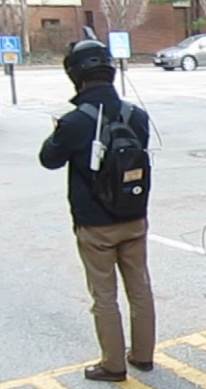}}
  \caption{A helmet network device. With this device, a human user can lead a robotic convoy team.}
  \label{fig:6-10}
\end{figure}

Depending on the application, a human user needs to lead the robotic convoy team. To do so, we developed a helmet device comprised of a network camera, AP with omni-directional antenna, battery, and network switch. This device was carried on the back of an operator as shown in Figures \ref{fig:6-11-a} and \ref{fig:6-11-b}, and the AP can properly transmit radio signals, which is detected by the receiver on the follower robot  and subsequently measured for WCA and to successfully track the human user. With the attached camera on a helmet, a human user can provide a command center with a live view of an interesting scene and communicate through a microphone on the camera. We do not present any real situations that require this device, but this can be used in a variety of applications such as for surveillance, search and rescue missions, etc. 

\subsection{Results}
\label{ch:6-6-2}
In order to validate the proposed system, three different test scenarios are conducted to evaluate 1) convoy strategy; 2) obstacle avoidance algorithm; 3) {\color{black}leader-follower robotic relay communication system}.

\begin{figure*}
\centering
\includegraphics[width=1.8\columnwidth]{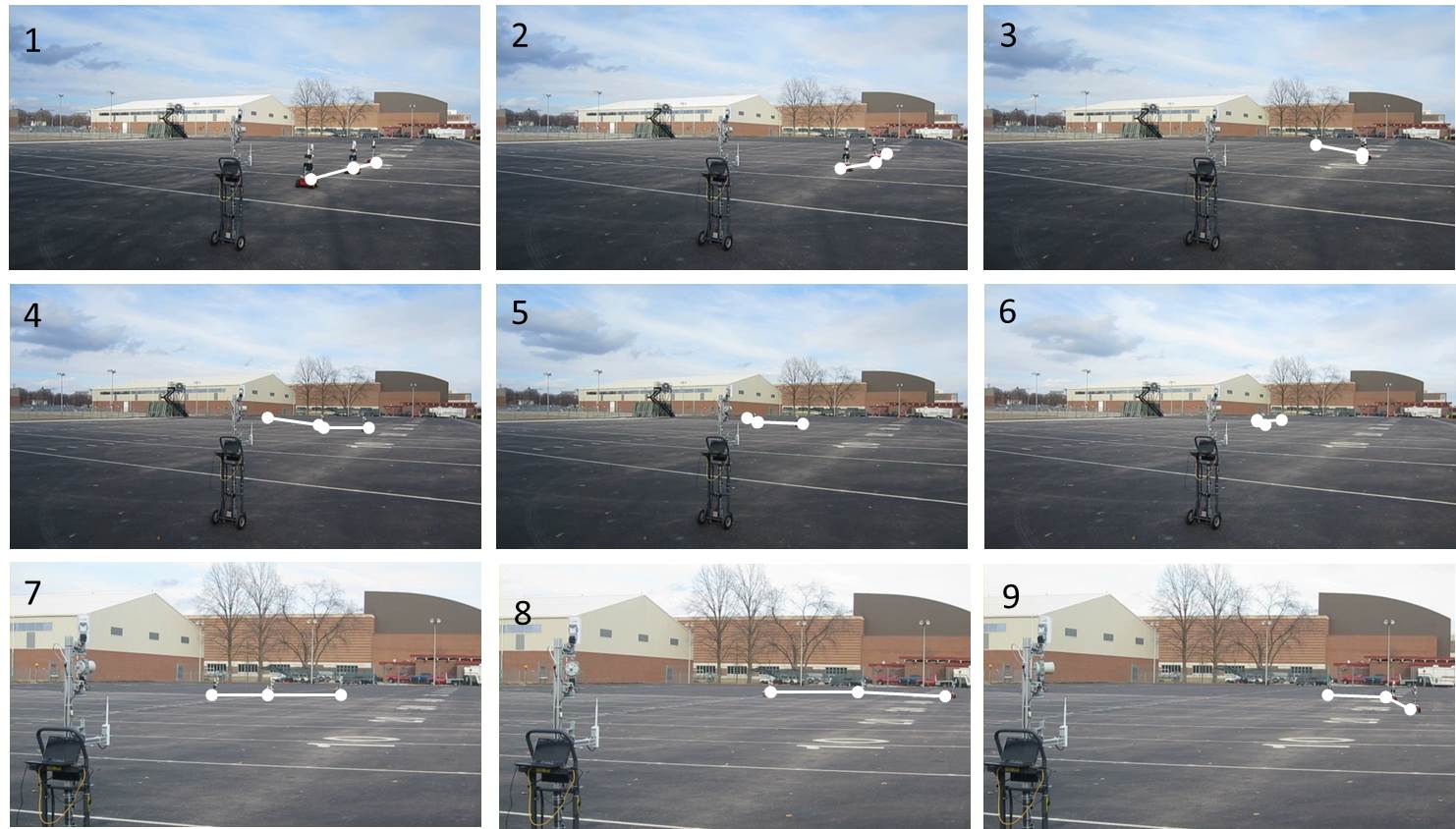}
\caption{Trace of robots' convoy in the field test with three robots, focusing on a robotic convoy strategy. A video demonstrating this field experiment can be found at \href{https://youtu.be/XymfE3qjeTc}{https://youtu.be/XymfE3qjeTc}.}
  \label{fig:6-11}
\end{figure*}

\subsubsection{Robotic Convoy Strategy}
\label{ch:6-6-2-1}
The first test set was designed to analyze the performance of the robot convoy strategy. Three robots were employed - one is a leader and the other two are followers. For this test, the leader robot was manually controlled with an averaged velocity set at 0.2 m/s, and the followers were initially placed in order behind the leader robot. ${Threshold}_l$ was set to $-30$ dBm for each follower so as to be closer to its leader and ${Threshold}_b$ to a large number, $-60$ dBm to be free from the distance constraint from a behind source, e.g., the last robot from the command center. Traces of all of the robots are depicted in Figure \ref{fig:6-11} with white round circles indicating the current positions of each robot at every 20 seconds. It clearly shows that every follower was able to properly track their leaders and thus successfully formed a convoy team.

\begin{figure}
  \centering 
  {\color{black}
  \subfigure[Traces of the leader and the follower]{\label{fig:enad2-a}  \includegraphics[width=0.60\columnwidth]{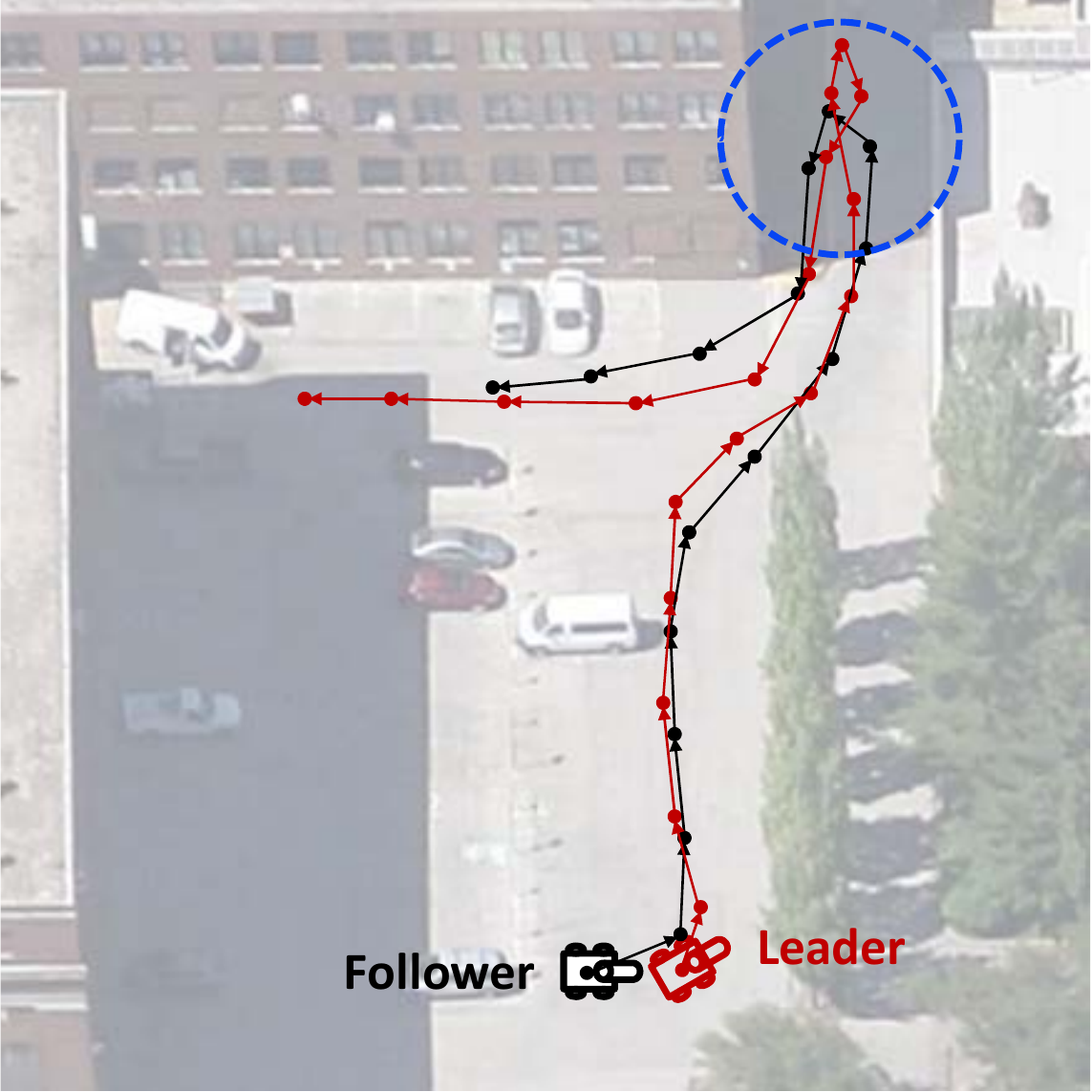}}%
  \hskip 0.5truein
  \subfigure[History of the best RSSI]{\label{fig:enad2-b}\includegraphics[width=0.95\columnwidth]{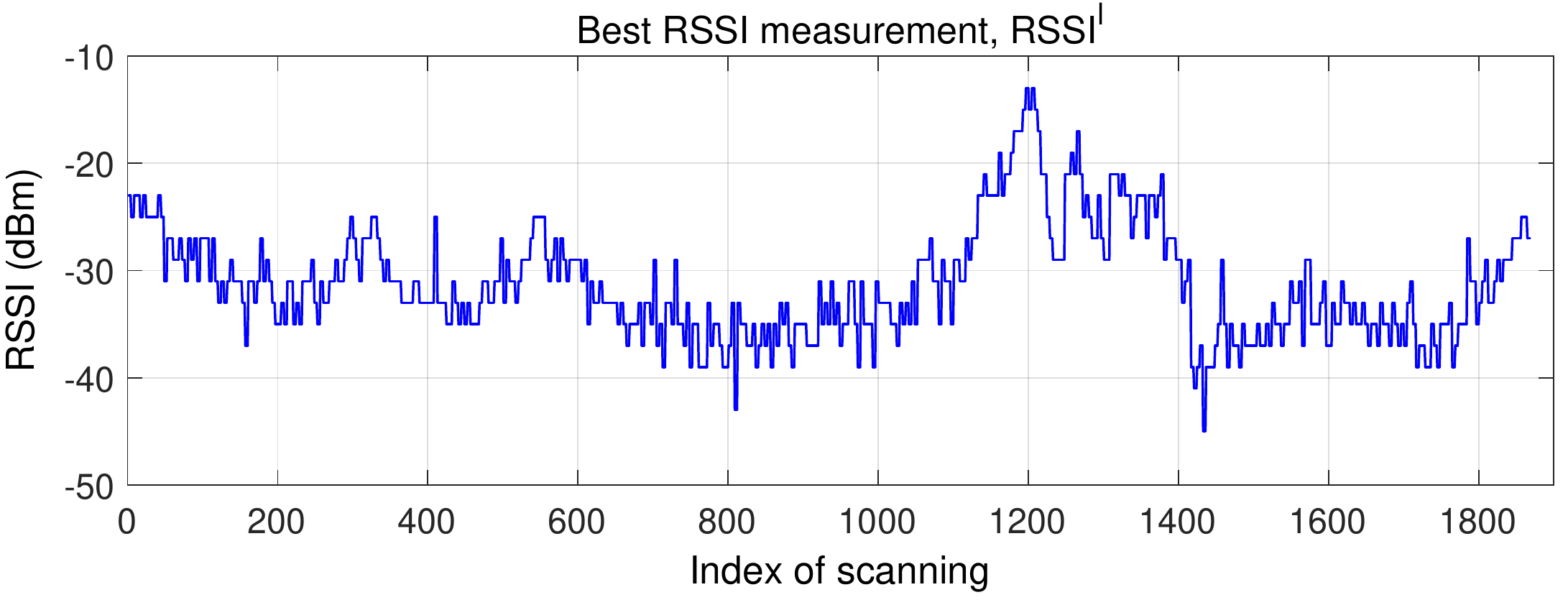}}
  \hskip 0.5truein
  \subfigure[History of the estimated bearing]{\label{fig:enad2-c}\includegraphics[width=0.95\columnwidth]{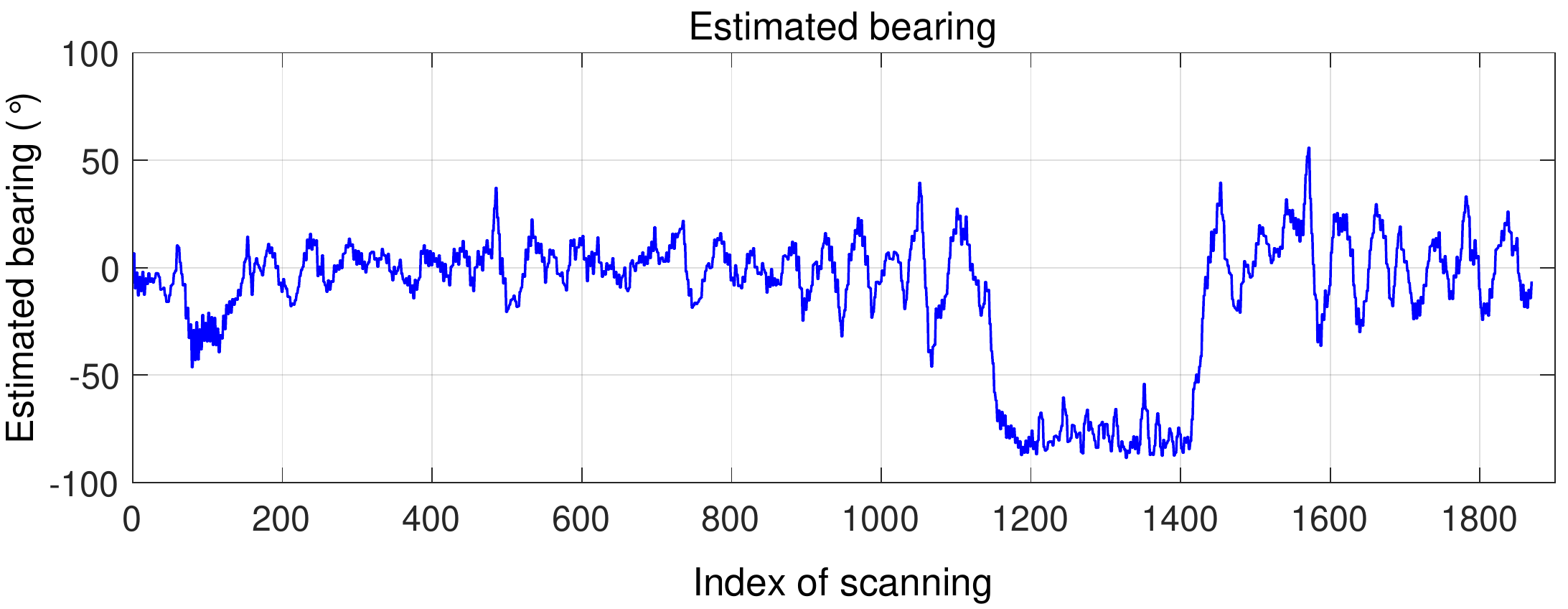}}
  \caption{%
  A field test focusing on a leader-follower control scheme. A video demonstrating this field experiment can be found at \href{https://youtu.be/1LFgR9naabU}{https://youtu.be/1LFgR9naabU}.}
  \label{fig:enad2}
  }
\end{figure}

{\color{black}Next, In order to analyze the performance of the leader-follower control scheme more in detail, we employed a pair of leader-follower robots. The leader robot was remotely controlled by a human operator at an averaged velocity of 0.2 m/s, and the follower was initially placed right behind the leader. For the stopping criteria, we set ${Threshold}_l$ to $-25$ dBm.

Figure \ref{fig:enad2} shows the results of this test. Figure \ref{fig:enad2-a} shows the tracked paths by the leader and the follower. The red lines depict movement paths by the leader, and the black lines depict movement paths by the follower. These lines were drawn by referring to videos recorded during the test and odometer information from the robots. This test also included multiple stops performed by the leader and very sharp paths requiring almost $180^\circ$ turning for the follower to successfully track the leader. As shown in Figure \ref{fig:enad2-a}, the follower followed the leader well during the entire test. 

Figure \ref{fig:enad2-b} shows that a history of the measured best RSSI denoted with ${RSSI}^l$. As shown in the horizontal axis, approximately 1900 scans were performed during this test. There were two times of stop taken by the leader during the first half of the test. As shown in Figure \ref{fig:enad2-b} around the 300\textsubscript{th} scanning and 500\textsubscript{th} scanning, the best RSSI reached the threshold accordingly, and therefore the follower stopped with a close distance to the leader. In addition, as shown in the right top of the map in Figure \ref{fig:enad2-a}, denoted with the blue dotted circle, the leader wheeled about to the other extreme. This was intended motion controlled by a human operator to see if the follower could follow the leader or not in such extreme situations. As the leader turned to the opposite direction that the follower headed, the estimated bearing by the reader showed all the way to the left (see around the 1200\textsubscript{th} scanning in \ref{fig:enad2-c}), meaning that the follower needs to change its heading to the left as well. However, since the measured RSSIs were higher than the threshold at that time, the follower had to stop for a while until the leader moved away from the follower (see around the 1200\textsubscript{th} scanning in Figure \ref{fig:enad2-b}. After the 1300\textsubscript{th} scanning, the best RSSI became lower than the threshold, and finally the follower wheeled round and resumed following the leader again. These behaviors validate that two robots in convoying perform well with the proposed leader-follower control scheme. }

\begin{figure}
  \centering 
  \subfigure[Traces of the two robots]{\label{fig:enad1-a}  \includegraphics[width=0.784\columnwidth]{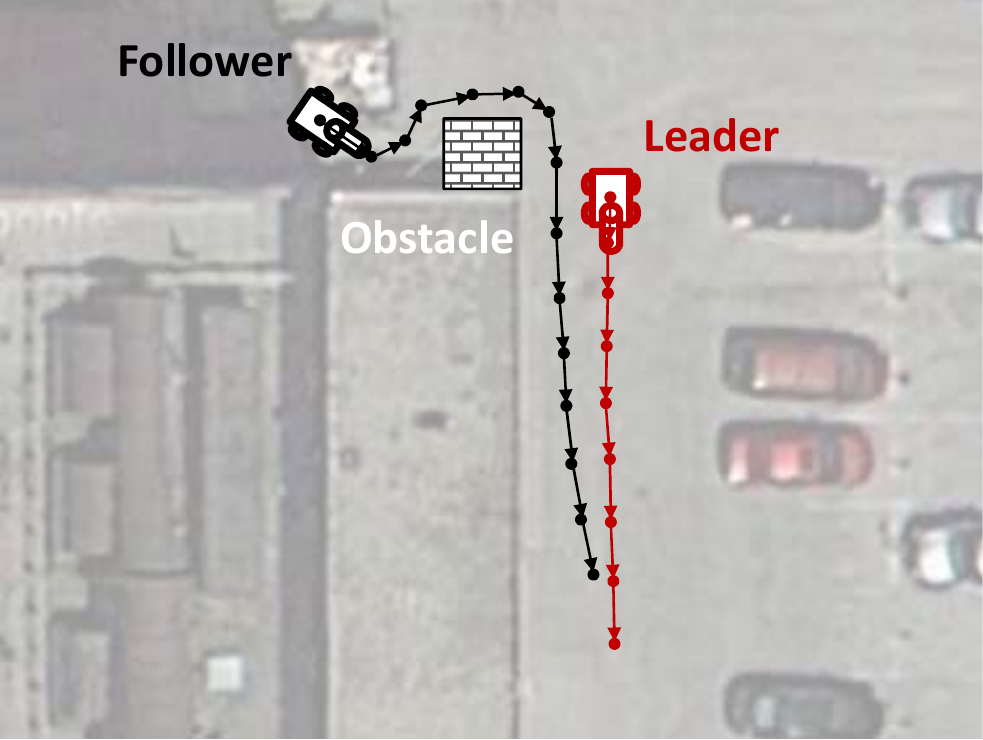}}
  \vskip 0.1truein
  \subfigure[History of the best RSSI]{\label{fig:enad1-b}\includegraphics[width=0.95\columnwidth]{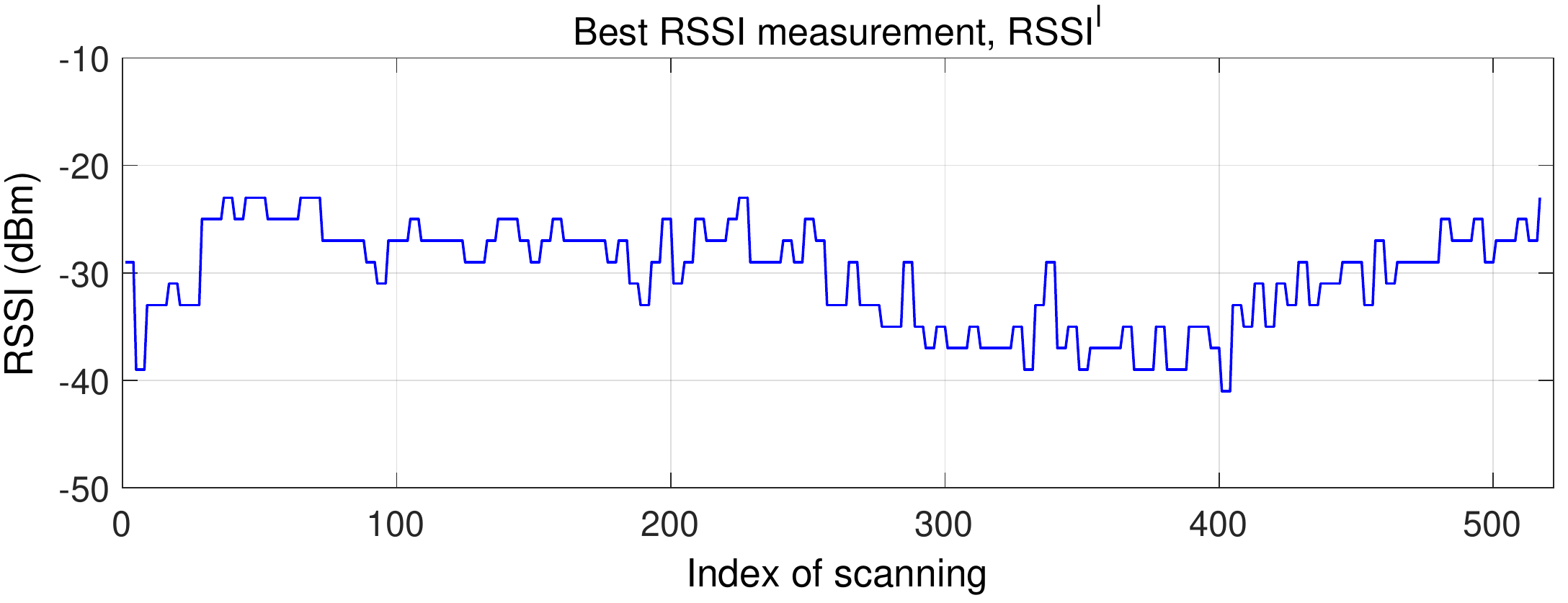}}
    \hskip 0.5truein
    {\color{black}
  \subfigure[History of the estimated bearing]{\label{fig:enad1-c}\includegraphics[width=0.95\columnwidth]{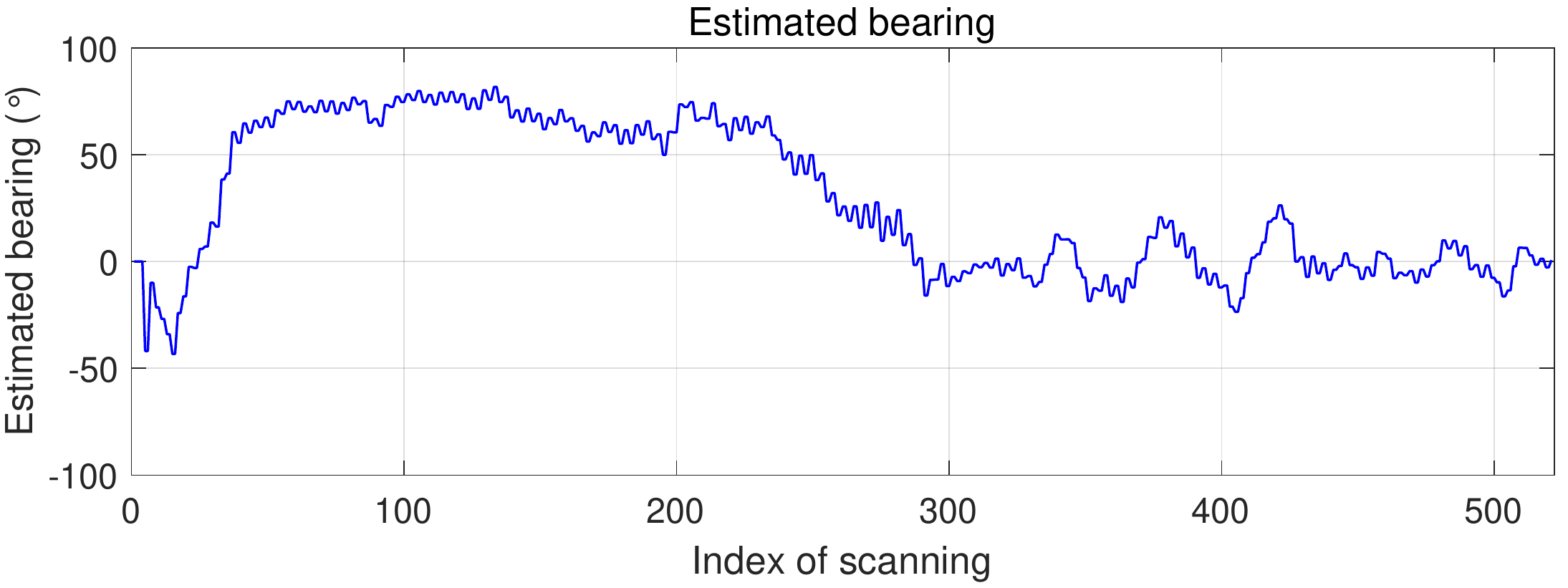}}
  }
  \caption{A field test focusing on an obstacle avoidance algorithm. A video demonstrating this field experiment can be found at \href{https://youtu.be/xyBxOtQcckc}{https://youtu.be/xyBxOtQcckc} and \href{https://youtu.be/YXoCh1MuQZE}{https://youtu.be/YXoCh1MuQZE}.}
  \label{fig:enad1}
\end{figure}

\subsubsection{Obstacle Avoidance Algorithm}
\label{ch:6-6-2-2}
In order to analyze the performance of the obstacle avoidance algorithm in detail, we {\color{black}again} employed a pair of leader-follower robots. The leader was manually controlled so that it moves straight to about 15 meters at a constant velocity of 0.2 m/s. The follower was initially placed behind the building and a square obstacle of an approximate size of $0.5 \times 0.5$ meters when viewed from above. In this planned situation, the follower should avoid the obstacle and the side of the building in order to successfully follow the leader. If otherwise, the follower fails to achieve its goal. For the stopping criteria, we set ${Threshold}_l$ to $-25$ dBm. 

In Figure \ref{fig:enad1-a}, the red lines indicate movement paths by the leader. The black lines indicate movement paths by the follower. These lines were drawn by referring to videos recorded during the test and odometer information from the robots. As shown in this figure, the follower was able to avoid the obstacle and the side of building without any contacts and {\color{black}to follow} the leader over the full {\color{black}paths}. Figure \ref{fig:enad1-b} shows a history of the measured best RSSI denoted with ${RSSI}^l$. As shown in the horizontal axis, approximately 520 scans were performed during this test. During the first half of scanning, there were few {\color{black}increases} in measured RSSI as the leader and the follower were close to each other. When the leader bore off gradually and the follower focused on avoiding obstacles, the measured RSSI decreased by up to about $-40$ dBm. However, as soon as the follower avoided the obstacles and {\color{black}became} free, it resumed following the leader. After that, as shown at the end of the history in Figure \ref{fig:enad1-b}, the measured RSSI reached the pre-defined threshold, $-25$ dBm, making the follower stop at a close distance to the leader.

{\color{black}
Figure \ref{fig:enad1-c} shows a history of the estimated bearings. As the leader departed from its original location, the estimated bearing by the follower showed very far to the right (see from around 40\textsubscript{th} scanning to around 200\textsubscript{th} in \ref{fig:enad2-c}), which is true considering the geometrical locations of the two robots (i.e., the leader was located in the far right side of the follower). However, the follower actually turned to the left direction, not to the right direction by the estimated bearing, to avoid the obstacle as shown in \ref{fig:enad2-a}. This was because the follower was in Situation 1. As soon as there is a change from Situation 1 to Situation 2, the robot resumed tracking the leader by integrating the sonar measurement into the antenna measurement. }

\subsubsection{{\color{black}Leader-follower Robotic Relay Communication System}}
\label{ch:6-6-2-3}

This subsection presents the results obtained from the experiments in complex indoor and outdoor environments. The results validate the robustness of the proposed leader-follower robotic relay communication system which combines the proposed convoy strategy and the obstacle avoidance algorithm.

\begin{figure}[t]
  \centering 
  \subfigure[Final locations of the user and relay nodes]{\label{fig:indoor-a}  \includegraphics[width=0.8\columnwidth]{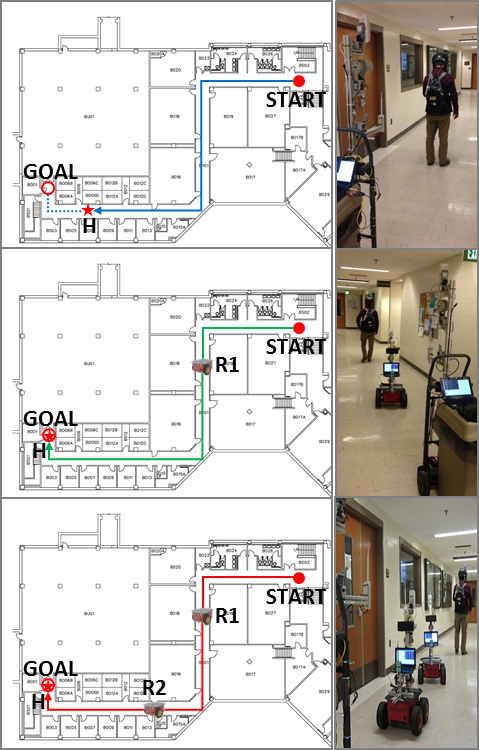}}
  \subfigure[Throughput measurement]{\label{fig:6-17}\includegraphics[width=0.95\columnwidth]{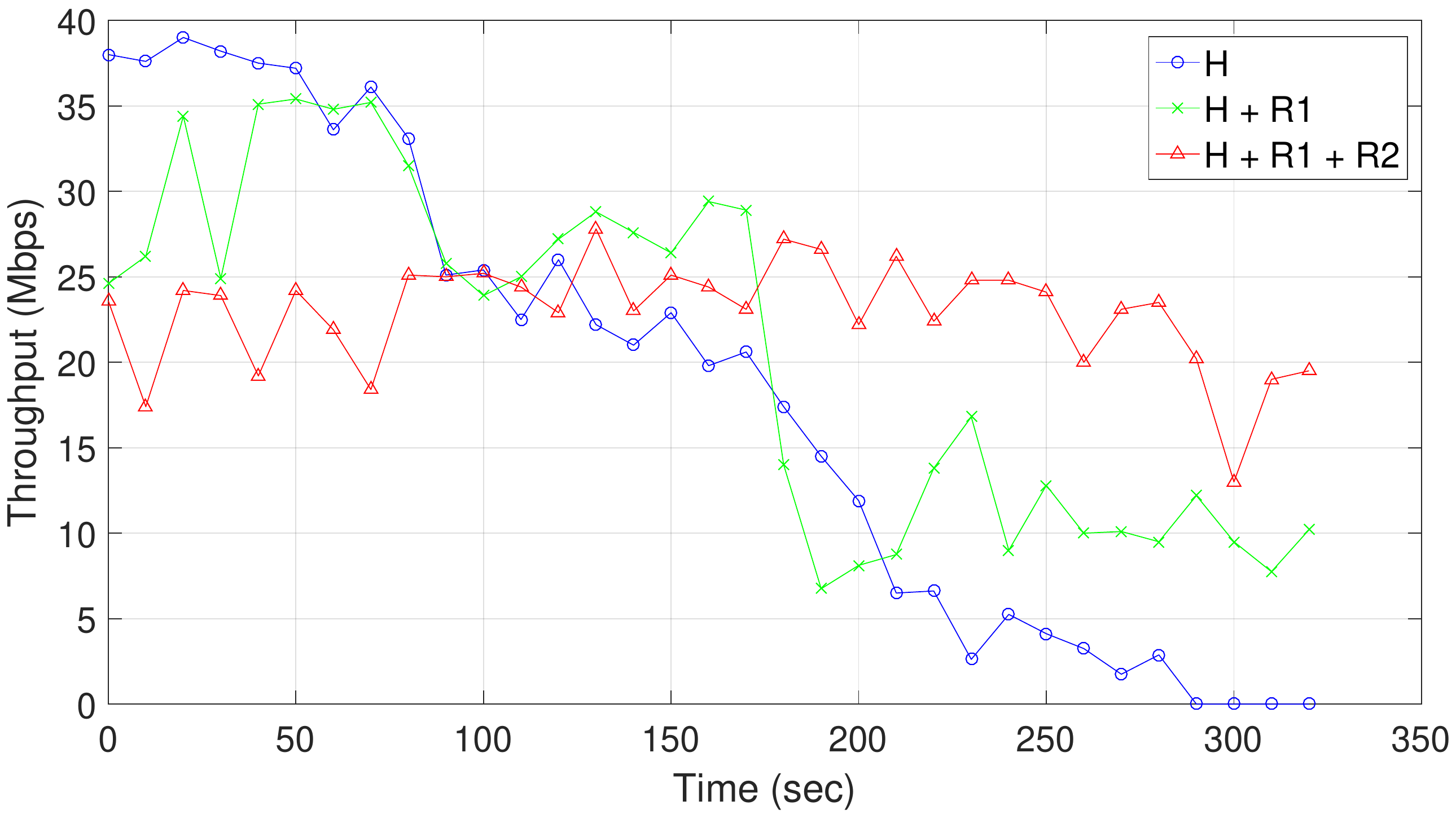}}
  \caption{Indoor tests for different numbers of relay nodes. A video demonstrating this field experiment can be found at \href{https://youtu.be/G5nWZR02nzM}{https://youtu.be/G5nWZR02nzM}; (a) A solid line indicates the actual trajectory of the end user which is represented as a red star ($\bigstar$). A solid circle in red ($\bullet$) and an empty circle in red ($\circ$) represent the staring point and the goal, respectively. Each figure shows the network connection range depending on the number of relay nodes. (b) End-to-end throughput measured at the end user in an indoor and complex environment.}
  \label{fig:indoor}
\end{figure}

First, an indoor test was conducted with three different sets as follows: 1) a human end user, 2) a human end user and a follower robot, and 3) a human end user and two follower robots. We conducted this test in Knoy Hall, one of the buildings at the Purdue main campus, because the long corridor of this building has both line-of-sight and non-line-of-sight regions and some invisible, but active wireless interference, making it a perfect place to test our proposed methods. During the test, we had the human user walk at a constant speed of 0.2 m/s. The designated path is depicted with a solid line in Figure \ref{fig:indoor-a}. ${Threshold}_l$ was set to $-30$ dBm for each follower to track its immediate leader and ${Threshold}_b$ to $-40$ dBm to maintain a connection with the node behind it. We observed: How far the human user could move away from the command center while maintaining high quality wireless connection. We used an end-to-end throughput measurement as a performance metric. 

In the results, the human failed to maintain a wireless connection with the command center with no relay robots before reaching the destination as shown in the top of Figure \ref{fig:indoor-a}. In this figure, the human user (denoted with ``H") almost reached to the destination (see the solid blue line indicating the path taken by the user), but could not reach the final destination (see the dashed line indicating an unvisited path by the user). This result could be expected since the test environment was composed of two sharp corners, placing the human in a non-line-of-sight region from the view of the command center, and was enclosed by thick brick walls with poor radio penetration. In comparison, the human could reach the destination with the aid of a relay robot (denoted with ``R1") and continue sending high-quality image data to the command center even at the final destination. The path taken is depicted in the middle of Figure \ref{fig:indoor-a} by a sold green line. In this experiment, the relay robot stayed at the initial position for a while, and then started tracking the trajectory of the human user after the RSSI threshold reached the designated value. During the most of the time the robot could not  see the human user due to non-line-of-sight, however, with the proposed bearing estimation, the robot was able to track the user successfully.  Subsequently, the robot passed around the first corner and stopped as the RSSI threshold from the node (i.e., the command center) behind it reached the designated value. The final location of the follower robot R1 is depicted in the middle of Figure \ref{fig:indoor-a}. 

Similarly, the human was also able to reach the destination with two relay robots (denoted with ``R1" and ``R2", respectively). In this setting, the first follower R2 (one behind the user) was able to track the human user until it passed the second corner, and the last follower R1 (one closest to the command center) started moving after a long time when the threshold from its leader reached the designated value. The final locations of the robots are depicted in the bottom of Figure \ref{fig:indoor-a}. As shown this figure, the last follower was located at almost same position as shown in the middle of Figure \ref{fig:indoor-a}, and the first follower passed two corners and was closer to the human user to maintain a wireless connection. It is worth noting that the both follower robots were able to track their front robot (i.e., a human user or a robot) with the proposed bearing estimation although most of the time they could not see it (non-line-of-sight).

The throughput measurements are summarized and depicted in Figure \ref{fig:6-17}. Initially, the first setting that employed no relay robots shows the highest throughput results, and the third setting that employed two relay robots shows the lowest results. In fact, this could be expected as described in \cite{pabst_relay-based_2004}. However, keeping in mind that our primary goal was to extend the radio range of the end node while maintaining good wireless connection. From the figure \ref{fig:6-17}, it can be seen that the throughput measured with the first setting decreased dramatically at around $80$ to $100$ seconds and $180$ to $200$ seconds. Around these time periods, the human passed the first and second corners and as such the wireless connection dramatically weakened, resulting in decreased throughput. Even worse, as the human approached the final destination, throughput went below 5 Mbps and finally the wireless network was disconnected (see after $280$ seconds). In contrast, when we tested the second setting exploiting a relay robot, a wireless network was able to stay connected to even the human user located in the final destination. The throughput result never went below 5 Mbps and reached around 10 Mbps at the end location with this setting, although there were two dramatic decreases in throughput measurements.  

Lastly, when we tested the third setting employing two relay robots and the wireless network maintained a very stable connection for the entire time period. More specifically, throughput was kept at around 20 to 25 Mbps after the human passed two corners, and was kept at around 20 Mbps even at the final location. This result validated that there was substantial range extension as a result of using the relay robot, and therefore this achieved our primary goal of extending the range of radio signals in indoor and complex environments. 

\begin{figure}[t]
  \centering 
  \subfigure[Final locations of the user and relay nodes]{\label{fig:outdoor-a}  \includegraphics[width=0.8\columnwidth]{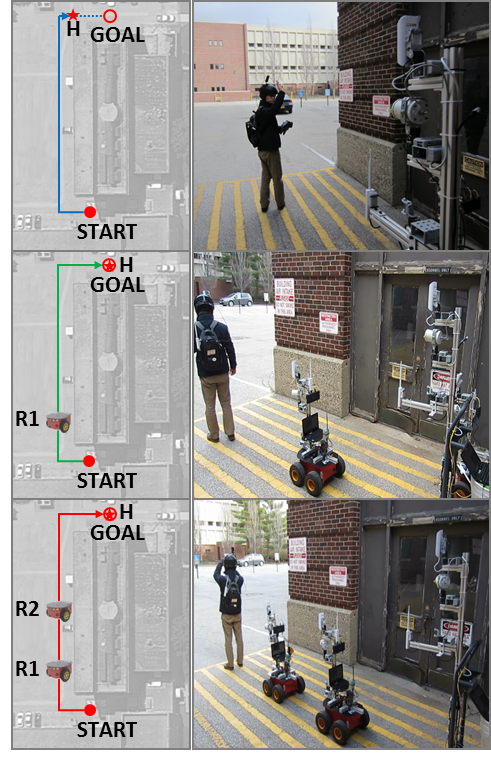}}
  \subfigure[Throughput measurement]{\label{fig:6-21}\includegraphics[width=0.95\columnwidth]{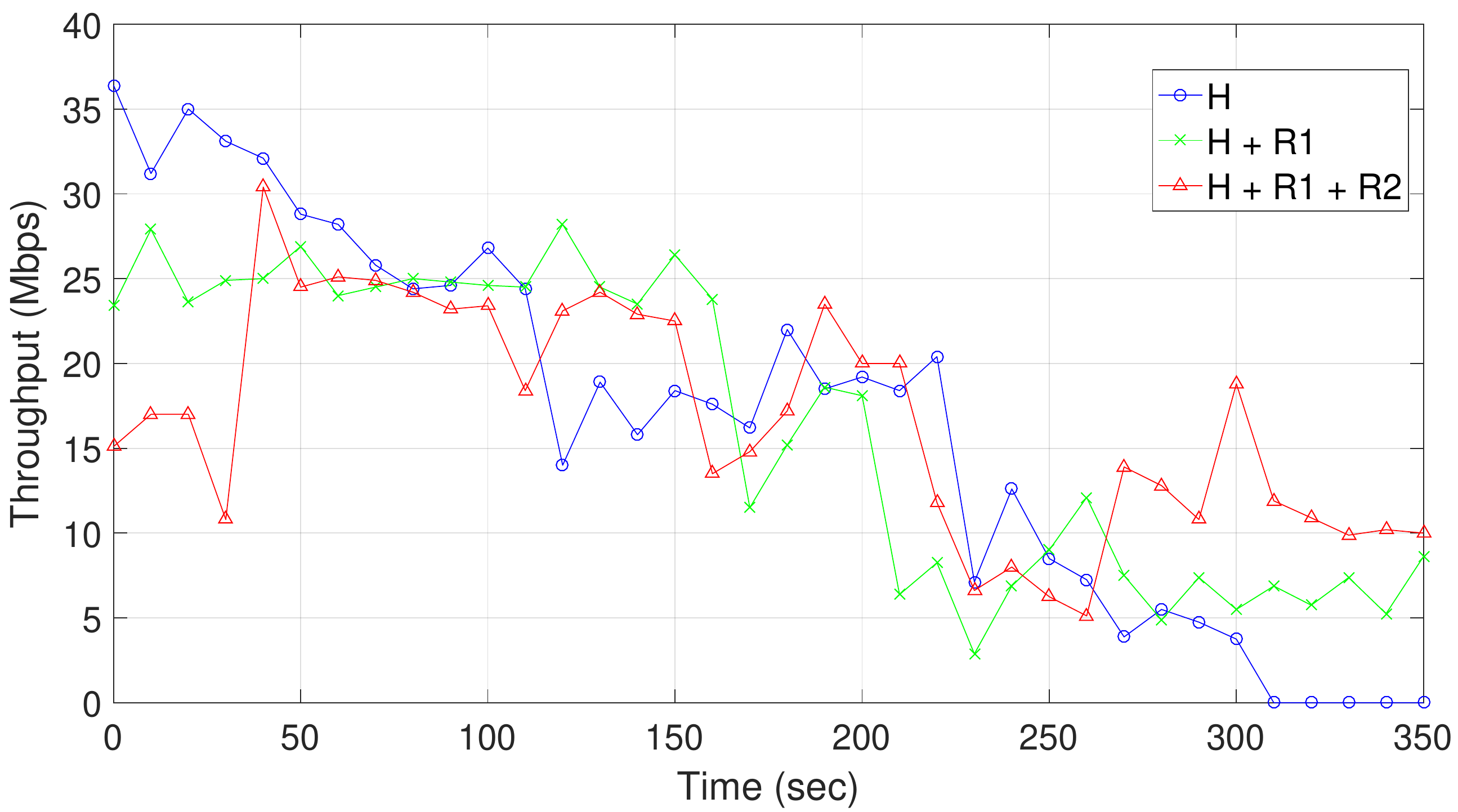}}
  \caption{Outdoor tests for different numbers of relay nodes. A video demonstrating this field experiment can be found at \href{https://youtu.be/lb3Oriv8Dvs}{https://youtu.be/lb3Oriv8Dvs}; (a) A solid line indicates the actual trajectory of the end user which is represented as a red star ($\bigstar$). A solid circle in red ($\bullet$) and an empty circle in red ($\circ$) represent the starting point and the goal, respectively. Each figure shows the network connection range depending on the number of relay nodes. (b) End-to-end throughput measured at the end user in an outdoor and complex environment.}
  \label{fig:outdoor}
\end{figure}

Second, an outdoor test was conducted with three different sets as follows: 1) a human end user, 2) a human end user and a follower robot, and 3) a human end user and two follower robots. For this test, we visited one of the parking lots at the Purdue main campus. This parking lot is large, with a $100 \times 70$ meter long area and thus requires a long-range radio communication if a command center and the designated final destination are placed at opposite ends of the environment. This environment also requires a non-line-of-sight service for a successful communication. The environment is shown in Figure \ref{fig:outdoor-a}. During the test, the human user walked at a constant speed of 0.2 m/s. The designated path is depicted with a solid line. We set the same thresholds as with the previous values.

This test obtained very similar results and trends as shown in the previous indoor test. First, with the first setting that employed no relay robots, the human (denoted with ``H") failed to stay connected to the command center before reaching the destination as shown in the top of Figure \ref{fig:outdoor-a}. Unlike the previous test in the indoor environment, the end user could be farther away from the command center, i.e., almost a total of 60 meters traveled. there appeared to be less radio interference and fewer effects on multipath propagation in this outdoor environment. However, as the total travel distance became longer, the human end user failed to pass around the second corner (see the dashed line in the top of Figure \ref{fig:outdoor-a}. 

In contrast, the other two settings successfully completed the mission with the human user reaching the destination while maintaining a quality of radio signal. In the second setting that exploited a relay robot R1, the robot tracked the trajectory of the human user well and stopped after passing the first corner by reaching the designated RSSI threshold with the command center. In the third setting that employed two relay robots, the first robot R2 (behind the human user) was able to keep tracking until it reached ${Threshold}_b$ from the robot behind it. Similarly, the last robot R1 kept tracking its leader (the first robot) well until it reached ${Threshold}_b$ from the node (i.e., command center) behind it.

Throughput measurements are summarized and shown in Figure \ref{fig:6-21}.  Similar to the previous test in the indoor environment, the first setting shows the highest throughput result, followed by the second setting, and the third setting shows the lowest result at the first 40 seconds. This trend was kept until the end node was in a line-of-sight region within  view of the command center. The trend soon changed after the human user turned around the first corner; that is, the three settings show very similar throughput data. However, as the end node approached the second corner, the throughput measured with the first setting went below 5 Mbps (see around $280$ seconds) and reached 0 Mbps (technically, no link) as the connection to the command center broke. In comparison, the other two settings were able to maintain a wireless connection with a command center at the final destination. Notably, the last selection was able to stay above 10 Mbps at the end, although there was a slight decrease in throughput measurements that appeared around $240$ to $260$ seconds.  
The results from this outdoor test validated the substantial range extension from using the relay robot, and therefore our primary goal to extend the range of radio signals in outdoor and complex environments was achieved.


\section{Conclusion and Future Work}\label{sec:conclusions}
In this paper, we proposed a robotic convoy system using directional antennas with the goal of creating end-to-end communication in dynamic and complex environments. With bearing estimation called WCA, directional antennas were utilized to guide a follower robot to its leader. As a result, our system yielded very robust direction estimations in a constrained environment, specifically when the follower was placed in a region out of line-of-sight with the leader. Furthermore, extensive field tests that included both  indoor and outdoor environments validated that our proposed method is fully decentralized and thus satisfies the scalability requirements of robotic network systems. By using the method, relay robots can ensure capacity and long-range end-to-end communication in robotic networks.

We considered simplified homogeneous disk models in order to model communication links between two neighboring nodes. Nonetheless, in realistic communication environments, each relay may need to have different models with respect to its surrounding environments. For example, a smaller disk size may need to be assigned to relays placed in a more dense space due to increased data loss that could be induced by effects of fading, shadowing and multipath. Conversely, for relays located in a more open space, a larger disk size would be more desirable. Therefore, adaptive models to take into account these issues should be considered so as to enhance network capability in end-to-end communications.

We do not consider situations where one of the radio links is broken and the links should be redefined as it is out of scope in this paper. Nonetheless for more practical use of the proposed system, we will research a backup plan or predefined action to cope with such situations. For example, a self-healing algorithm for mobile wireless sensor network will be investigated \cite{younis2008localized, senouci2014localized, izadi2015alternative}. 

\bibliographystyle{IEEEtran}
\bibliography{mybibfile}

\end{document}